\documentclass[12pt]{elsarticle}
\usepackage[algoruled,ruled]{algorithm2e}
\SetKwIF{If}{ElseIf}{Else}{if}{}{else if}{else}{end if}%
\usepackage{algpseudocode}
\usepackage{amsmath,amssymb,amsfonts,amsthm}
\usepackage{bm}
\usepackage{booktabs}
\usepackage{caption}
\usepackage{mathtools}
\usepackage{multirow}
\usepackage[T1]{fontenc}
\usepackage[numbers]{natbib}
\usepackage{siunitx}
\usepackage{soul}
\usepackage{subfig}
\usepackage{textcomp}
\usepackage{threeparttable}
\usepackage{makecell}
\usepackage{geometry}
\usepackage[color=green!40]{todonotes}
\soulregister\cite7
\soulregister\footnote7
\soulregister\eqref7
\soulregister\ref7
\usepackage{hyperref}
\newcommand{\bigO}{\mathcal{O}}

\makeatletter
\def\ps@pprintTitle{%
  \let\@oddhead\@empty
  \let\@evenhead\@empty
  \let\@oddfoot\@empty
  \let\@evenfoot\@oddfoot
}
\makeatother

\graphicspath{{./}{fig/}{../fig/}}

\begin{document}

\begin{frontmatter}
  \title{Mean-Teacher-Based Semi-Supervised Learning Framework for
    Scalable Indoor Localization Using Wi-Fi RSSI Fingerprinting\tnotemark[1]}
  \author[siit,xjtlu]{Sihao Li}
  \ead{01177@siit.edu.cn}
  \author[zut,xjtlu]{Zhe Tang}
  \ead{tangzhe@bcszjut.com}
  \author[xjtlu]{Kyeong Soo Kim\corref{cor1}}
  \ead{Kyeongsoo.Kim@xjtlu.edu.cn}
  \author[uol]{Jeremy S. Smith}
  \ead{J.S.Smith@liverpool.ac.uk}
  \affiliation[siit]{%
    organization={School of Artificial Intelligence},%
    addressline={Suzhou Institute of Industrial Technology},%
    city={Suzhou},%
    postcode={215104},%
    state={Jiangsu},%
    country={China}%
  }%
  \affiliation[zut]{%
    organization={Institute of Artificial Intelligence Innovation},%
    addressline={Zhejiang University of Technology},%
    city={Hangzhou},%
    postcode={310000},%
    state={Zhejiang},%
    country={China}%
  }%
  \affiliation[xjtlu]{%
    organization={School of Advanced Technology},%
    addressline={Xi'an Jiaotong-Liverpool University},%
    city={Suzhou},%
    postcode={215123},%
    state={Jiangsu},%
    country={China}%
  }%
  \affiliation[uol]{%
    organization={Department of Electrical Engineering and Electronics},%
    addressline={University of Liverpool},%
    city={Liverpool},%
    postcode={L69 3GJ},%
    country={U.K.}
  }%

  \cortext[cor1]{Corresponding author}
  \tnotetext[1]{%
    This work is supported in part by the Research Foundation for Advanced
    Talents of Suzhou Institute of Industrial Technology (under Grant SIIT
    2026KYQD006) and the Postgraduate Research Scholarships (under Grant
    PGRS1912001) and the Key Program Special Fund (under Grant KSF-E-25) of
    Xi'an Jiaotong-Liverpool University.}
  \begin{abstract}
    Conventional large-scale indoor localization based on Wi-Fi RSSI fingerprinting
    faces issues of time-consuming and labor-intensive labeled data collection,
    limited generalization of a model trained under a supervised learning (SL)
    framework due to its inability to leverage unlabeled data, and model
    performance degradation in dynamic scenarios with environmental variations. To
    address those challenging issues, we propose a comprehensive semi-supervised
    learning (SSL) framework for a deep neural network (DNN) localization model
    based on the Mean Teacher, which incorporates access point selection, model
    pre-training/cloning, and batch-level noise injection. The proposed SSL
    framework can not only efficiently use hybrid labeled/unlabeled databases for
    static training of a model during the offline phase, but also exploit unlabeled
    fingerprints from users of the indoor localization system deployed in the field
    for continuous retraining of the model during the online phase. We base the
    proposed SSL framework on the Mean Teacher because it can generate more stable
    target labels through an exponential moving average of model weights without
    incurring the high computational complexity of the Pi-Model and with better
    scalability for online learning than Temporal Ensembling, making it an optimal
    choice that strikes the right balance between performance and computational
    complexity in large-scale indoor localization. With the UJIIndoorLoc database,
    the proposed SSL framework reduces the mean 3D errors of the CNNLoc and
    SIMO-DNN models by 7.403\% and 7.748\%, respectively, compared with those under
    the conventional SL framework; with the XJTLU dynamic database, the maximum
    reduction in mean 2D error reaches up to 49.227\% under a dynamic training
    scenario, demonstrating the substantial performance improvement achieved by the
    proposed SSL framework.
  \end{abstract}



\end{frontmatter}

\section{Introduction}
\label{sec:intro}
As people spend increasing amounts of time in places like shopping malls,
airports, and hospitals, there has been an increasing need for indoor
location-based services (LBS)~\cite{intor_01_1,intor_01_2}. However,
localization systems based on the popular global navigation satellite system
(GNSS) are inadequate for indoor use due to the lack of line-of-sight (LOS)
paths to GNSS satellites~\cite{intro_02}.

As an alternative indoor LBS technology that does not rely on
satellites~\cite{intro:survey_03,intro:survey_04}, Wi-Fi received signal
strength indicator (RSSI) fingerprinting has become the most popular technique
thanks to the ubiquity of Wi-Fi infrastructure indoors. However, due to complex
multipath fading indoors, Wi-Fi RSSIs cannot be directly used for distance
estimation based on a path loss model, which is the fundamental principle of
multilateration~\cite{intro:survey_05}. Instead, Wi-Fi RSSI fingerprinting uses
RSSIs measured at a given location as a location fingerprint, i.e., a unique
pattern corresponding to each location, to determine the location based on
classification or regression. In Wi-Fi RSSI fingerprinting, first, a
fingerprint database is constructed based on RSSIs from all access points (APs)
measured at reference points (RPs), which are known locations specified by
their two-dimensional (2D)/three-dimensional (3D) coordinates and/or location
labels like room numbers; the constructed fingerprint database is then used to
train an indoor localization model during the offline phase. Finally, the
trained model is used to estimate the current location of a user or a device
based on the measured RSSIs during the online (real-time)
phase~\cite{intro_02}.

Since traditional Wi-Fi fingerprinting indoor localization approaches rely on
conventional machine learning (ML) algorithms, e.g., k-nearest neighbors (KNN)
and weighted k-nearest neighbors (wKNN)~\cite{knn_01,knn_02}, they require a
substantial amount of manual parameter tuning and, therefore, are not suitable
for indoor localization in complex, multi-building, and multi-floor
environments. Instead, researchers in this field increasingly focus on deep
learning (DL)-based techniques for large-scale indoor localization. To address
the scalability issues in large-scale multi-building and multi-floor indoor
localization, deep neural network (DNN) models have been proposed to enable
scalable indoor localization~\cite{Kim:18-1,Kim:18-3,cnn_01} using supervised
learning (SL) algorithms. In large-scale indoor localization scenarios (e.g.,
the UJIIndoorLoc database covering multiple floors of three buildings),
collecting and annotating labeled data incur substantial labor and time costs,
and dynamic changes in the wireless environment necessitate continuous model
updates. Due to their inability to utilize unlabeled data from the user side,
SL frameworks suffer from high model maintenance costs and limited
generalization ability; in contrast, SSL technology can efficiently utilize
mixed labeled and unlabeled data, which not only reduces the cost of labeled
data collection during the offline phase but also enables dynamic model updates
through unlabeled data during the online phase, thereby serving as a key
technical support for large-scale and long-term indoor localization deployment.

\begin{figure}[!htb]
  \centering%
  \includegraphics[angle=-90,width=.8\textwidth]{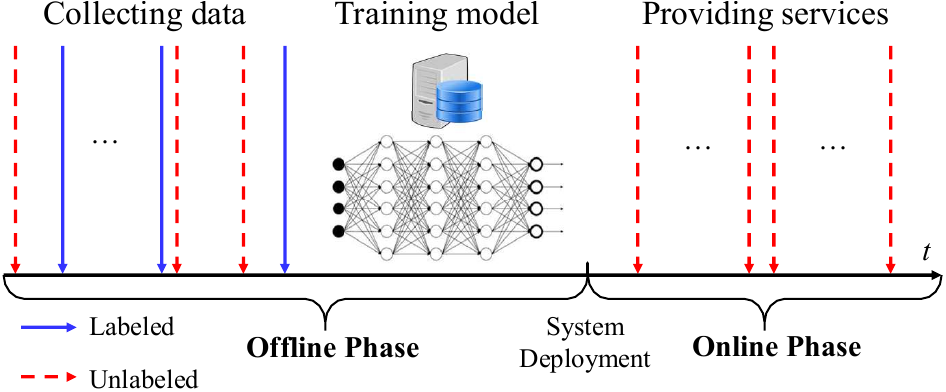}
  \caption{A real-world scenario for an indoor localization system based on
    Wi-Fi RSSI fingerprinting, where unlabeled fingerprint data are provided to
    the system.}
  \label{fig:ssl_sce}
\end{figure}

Fig.~\ref{fig:ssl_sce} shows a real-world scenario for an indoor localization
system based on Wi-Fi fingerprinting, where unlabeled fingerprint data are
provided to the system.

In this scenario, the unlabeled data available during the offline phase (e.g.,
from crowdsourcing volunteers) cannot be utilized for SL training unless
manually labeled. Likewise, the significant amount of unlabeled data generated
by system users during the online phase cannot be leveraged to update the
localization model. Consequently, the quality of the indoor localization
service may degrade progressively over time.

Given the limitations and challenges of the conventional SL framework,
researchers have investigated indoor localization based on semi-supervised
learning (SSL) to exploit unlabeled data. Among SSL methods,
\textit{consistency regularization} has emerged as a prominent technique. It
encourages a model to make consistent predictions on labeled and unlabeled
data, resulting in improved model generalization. Under the category of
consistency regularization, the Mean Teacher~\cite{ssl:mean_teacher} is a
practical method that has gained popularity in various applications. However,
its application to indoor localization based on Wi-Fi RSSI fingerprinting has
yet to be fully investigated.

Therefore, in this paper, we propose a comprehensive SSL framework for DNN
models for scalable indoor localization, which incorporates AP selection, model
pre-training/cloning, batch-level noise injection, and Mean Teacher-based SSL
training. We evaluate the superiority of the proposed SSL framework over the
conventional SL framework for the following two representative application
scenarios based on the real-world scenario shown in Fig.~\ref{fig:ssl_sce}:
\begin{itemize}
  \item \textit{Static training using hybrid labeled/unlabeled databases}:
        A localization model is trained using a hybrid labeled/unlabeled fingerprint
        database constructed through in-house/outsourced collection efforts and
        voluntary contributions during the offline phase.
  \item \textit{Dynamic training during the online phase}:
        The localization model, trained during the offline phase, is continuously
        retrained based on unlabeled data from users of an indoor localization system
        deployed in the field during the online phase to enhance its long-term
        performance.
\end{itemize}

The rest of the paper is structured as follows: Section~\ref{sec:rw} reviews
the existing work on indoor localization methods based on SL and SSL.
Section~\ref{sec:ssl} presents the proposed SSL framework for scalable indoor
localization. Section~\ref{sec:exp-results} presents the experimental results
based on the UJIIndoorLoc multi-building and multi-floor database and the XJTLU
dynamic database for the two SSL scenarios of static and dynamic training.
Finally, Section~\ref{sec:con} concludes our work in this paper.

\section{Related Work}
\label{sec:rw}
\subsection{DNN-Based Indoor Localization Models}
As the first DNN-based model for multi-building and multi-floor indoor
localization, the scalable deep neural network (DNN) architecture is proposed
in~\cite{Kim:18-1} to address the scalability issues in large-scale indoor
localization, where the number of output nodes is significantly reduced by
changing the location estimation framework from multi-class classification to
multi-label classification. The single-input and multi-output DNN (SIMO-DNN)
architecture proposed in~\cite{Kim:18-3} further enhances the scalable DNN
design by allowing flexible training of building, floor, and location outputs
and their dedicated hidden layers using different estimation frameworks; for
instance, building and floor estimations are based on multi-class
classification, while floor-level location estimation is based on the
regression of 2D coordinates. Therefore, the number of output nodes for the
floor-level location estimation is greatly reduced from the maximum number of
RPs per floor to two thanks to the use of the regression framework.

In~\cite{cnn_01}, the integration of stacked autoencoders (SAEs) and
convolutional neural networks (CNNs) is investigated to create a hybrid model
called CNNLoc. This model is comprised of three networks for estimating
building, floor, and 2D locations. The SAE is connected to fully-connected
layers featuring three-output nodes dedicated to building classification. To
classify floors, a dropout layer is introduced in the SAE, which then passes
features to a one-dimensional CNN (1D-CNN) that includes a fully-connected
layer as the output, resulting in five outputs for one-hot-encoded floor
classification. With some modifications, a similar 1D-CNN structure is employed
for the position estimator, adjusting the number of output nodes from five to
two to represent longitude and latitude coordinates. CNNLoc~\cite{cnn_01} is
recognized as one of the most classical CNN-based models for multi-building and
multi-floor indoor localization and is frequently used as a benchmark in the
literature.

\subsection{Semi-Supervised Learning Methods}
SSL methods can be categorized into the following four groups: Consistency
regularization, proxy label methods, generative models, and graph-based
methods~\cite{ssl:overview1}. Of these groups, in this paper, we focus on
consistency regularization as an SSL framework for large-scale indoor
localization for the following reasons:
\begin{itemize}
  \item Proxy label methods rely on the quality of pseudo-labels, which are used as
        labels; when a model cannot predict high-quality pseudo-labels, its performance
        will be significantly degraded.
  \item Generative models and graph-based methods are difficult to train and tune due
        to the complexity of the models, and, as a result, their performance is not
        always guaranteed.
\end{itemize}

The Pi-Model~\cite{ssl:temporal_ensembling}, one of the earliest SSL models
based on consistency regularization, is trained to minimize the difference
between its predictions on the labeled and unlabeled data. The idea is that by
maximizing the entropy, the model is encouraged to make more diverse
predictions on the unlabeled data with an expectation to improve the model's
generalization capability. A major issue is its higher computational complexity
resulting from the requirement of two forward propagation steps per input to
compute the consistency loss.

To address the complexity issue of the Pi-Model, Temporal
Ensembling~\cite{ssl:temporal_ensembling} is proposed that can reduce the
number of forward propagation steps per input to one. Temporal Ensembling
involves forming a consensus prediction of the unknown labels using the outputs
of the network-in-training on different epochs, different regularization, and
input augmentation conditions. This ensemble prediction can be expected to be a
better predictor for the unknown labels than the network output at the most
recent training epoch and thus can be used as a target for training.

The Mean Teacher~\cite{ssl:mean_teacher} is an SSL method based on averaging
model weights instead of predictions, which can provide two practical
advantages of (1) more accurate target labels by a teacher model and (2) better
scalability to large datasets and online learning over Temporal Ensembling.
Unlike Temporal Ensembling, the Mean Teacher updates the weights of the teacher
model based on exponential moving averages (EMAs) of the weights of a student
model and uses them to generate the teacher model's predictions at each
training step. During training, the Mean Teacher adjusts the weights of the
student model to make predictions closer to those of the teacher model; when
the predictions of the teacher model are ambiguous, the Mean Teacher allows the
student model to make its predictions.

\subsection{Indoor Localization Based on Semi-Supervised Learning}
Although SSL has been widely applied to various fields, its application to
indoor localization based on Wi-Fi fingerprinting is still in an early stage.

One of the earliest works is the graph-based SSL (GSSL) method proposed by
Zhang et al.~\cite{rela_GSSL}, which exploits the correlation between RSS
values at nearby locations to estimate optimal RSS values and improves the
smoothness of the radio map and localization accuracy for indoor localization
based on crowdsourced data. In~\cite{rela_TSLSSL}, a time-series SSL algorithm
is proposed for indoor localization, where unlabeled data are utilized to
generate pseudo-labels to improve the efficiency and accuracy of the
positioning model based on RSSI measurements.

The adapted mean teacher (AMT) model~\cite{rela_MT_CIR} introduces indoor
fingerprint positioning based on the channel impulse response (CIR), while Chen
et al.~\cite{rela_SSLComp} compares the performances of SSL-based channel state
information (CSI) fingerprinting techniques using variational autoencoder (VAE)
and generative adversarial network (GAN) models.

In~\cite{rela_WGAN}, the authors propose two approaches to address the problem
of limited labeled data in indoor localization: The first approach is a
weighted semi-supervised DNN-based method that combines labeled samples with
inexpensive pseudo-labeled samples to improve localization accuracy. The second
approach is a weighted GAN-based method that generates fake fingerprints to
overcome the unavailability of unlabeled data.

In addition to those papers focusing on SSL models, SSL algorithms are applied
to solve indoor localization problems at a framework level.
WePos~\cite{rela_WePos} and MTLoc~\cite{rela_MTLoc} are two representative
studies based on Wi-Fi RSSI fingerprinting: The WePos is based on a weak
supervision framework founded on bidirectional encoder representations from
transformers (BERT)~\cite{rela_BERT}, using weakly-labeled data to tackle
indoor localization on a large scale. The experiments are conducted in a vast
shopping mall. The MTLoc introduces a multi-target domain adaptation network
(MTDAN) that uses labeled and unlabeled data to improve the generalization
capability of a model. The MTLoc prioritizes feature extraction, utilizing a
complex GAN structure to extract long-term stable or semi-stable APs.

\subsection{Limitations of Existing Works Based on Semi-Supervised Learning}
Compared to the proposed Mean Teacher-based SSL framework, existing SSL methods
for indoor localization exhibit limitations in adaptability, computational
complexity, and scalability:
\begin{enumerate}
  \item \textbf{Graph-based SSL:}
        This category (e.g.,~\cite{rela_GSSL}) relies on the spatial correlation of
        RSSI to construct graph structures, resulting in high computational complexity
        and incompatibility with large-scale multi-building and multi-floor scenarios.
  \item \textbf{GAN/VAE-based SSL:}
        These methods (e.g.,~\cite{rela_SSLComp, rela_WGAN, rela_MTLoc}) require
        designing complex generators and discriminators, exhibit difficulty in
        hyperparameter tuning, and are prone to generating false fingerprint data
        without physical significance; consequently, localization accuracy could be
        deteriorated by false synthetic data.
  \item \textbf{Weakly Supervised SSL:}
        Methods like WePos~\cite{rela_WePos}, built on BERT, require numerous
        parameters, making deployment on mobile or edge devices challenging and
        preventing support for online dynamic training.
\end{enumerate}
In contrast, the proposed SSL framework adopts a simple and efficient
teacher-student structure, requiring the deployment of only the teacher model on
user devices with low computational overhead. It does not require complex
generator design and can generate unlabeled data consistent with real-world
characteristics through batch-level noise injection. Additionally, it supports
online dynamic training, enhancing the long-term performance of deployed indoor
localization systems.

\section{Semi-Supervised Learning Framework for Scalable Indoor Localization}
\label{sec:ssl}
\begin{figure}[!htb]
  \centering%
  \includegraphics[angle=-90,width=.75\textwidth]{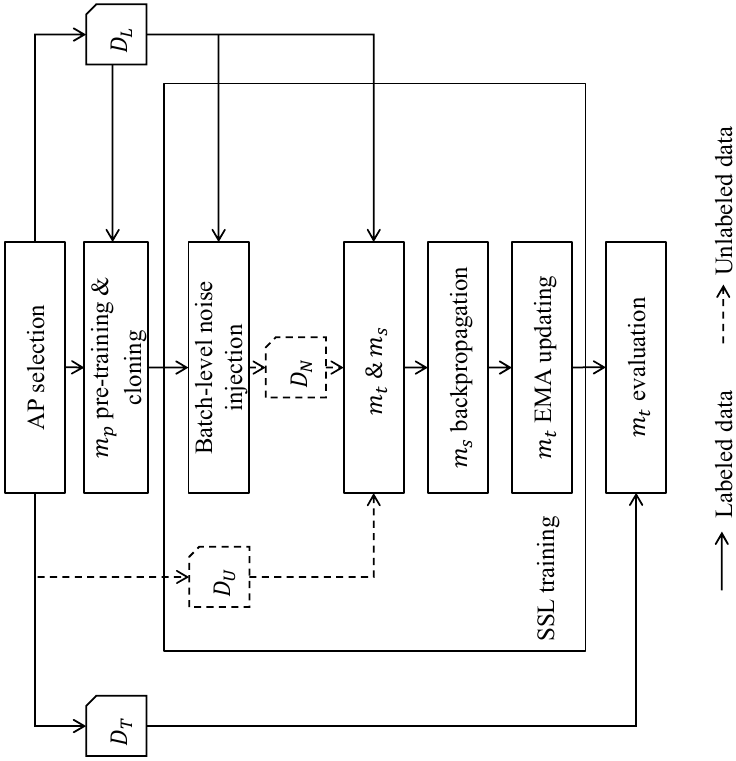}
  \caption{An overview of the proposed SSL framework for scalable indoor
    localization.}
  \label{fig:ts_alg}
\end{figure}

Fig.~\ref{fig:ts_alg} shows an overview of the SSL framework that we propose
for scalable indoor localization based on DNN models, utilizing the consistency
regularization strategy of the Mean Teacher~\cite{ssl:mean_teacher}.

The proposed SSL framework comprises three main components of AP selection,
$m_{p}$ pre-training and cloning, and SSL training, where $m_{p}$, $m_{s}$, and
$m_{t}$ respectively denote pre-training, student, and teacher models, all of
which are based on the same DNN architecture. Two groups of labeled, $D_{L}$
and $D_{T}$, and unlabeled, $D_{U}$ and $D_{N}$, datasets are used for the
proposed framework. $D_{L}$, $D_{T}$, $D_{U}$, and $D_{N}$ represent labeled,
test, unlabeled, and noise-injected datasets, respectively, and can be
formulated as follows:
\begin{equation}
  \begin{aligned}
    D_{L} & = \left\{\left(\bm{x}_{L,i},\bm{y}_{L,i}\right) | i = 1, \ldots,
    n_{L}\right\},                                                           \\ D_{T} & = \left\{\left(\bm{x}_{T,i},\bm{y}_{T,i}\right) | i =
    1, \ldots, n_{T}\right\},                                                \\ D_{U} & = \left\{\bm{x}_{U,i} | i = 1, \ldots,
    n_{U}\right\},                                                           \\ D_{N} & = \left\{\bm{x}_{N,i} | i = 1, \ldots, n_{N}\right\},
  \end{aligned}
  \label{eq:datasets}
\end{equation}
where $x_{L,i}$, $x_{T,i}$, $x_{U,i}$, and $x_{N,i}$ represent features (i.e.,
RSSIs), and where $y_{L,i}$ and $y_{T,i}$ represent ground truths (i.e.,
location coordinates or a label), respectively.

\subsection{AP Selection}
\label{sec:ap-selection}
We first perform feature selection based on APs where the APs providing RSSIs
which contribute the most to the localization are selected, which can not only
reduce the input dimensionality but also enhance the generalization capability
of the subsequent training in the proposed framework. The AP selection can also
provide a more stable, informative, and well-aligned input for long-term indoor
localization services (e.g., eliminating temporary hotspots).

Although there have been suggested AP selection schemes based on the
statistical characteristics of RSSIs
(e.g.,~\cite{AP_SLC_02,AP_SLC_03,AP_SLC_04,AP_SLC_05}), most of them are based
on labeled data or require complex processes. However, in the context of SSL,
where the primary goal is to ensure feature consistency between labeled and
unlabeled data, such strict selection criteria should be relaxed and revised.

Therefore, in the proposed framework, we select APs based on their numbers of
unique RSSI values~\cite{AP_SLC_01} as described in
Algorithm~\ref{alg:ap-selection}, where the threshold $\tau$ is a
hyperparameter that can be tuned for a given database.
\begin{algorithm}[!htb]
  \caption{AP selection.}
  \label{alg:ap-selection}
  \SetAlgoLined%
  \DontPrintSemicolon%
  \SetKwInOut{Input}{Input}%
  \SetKwInOut{Output}{Output}%
  \Input{Original fingerprint database $\mathcal{D}_{org}$.}%
  \Input{Threshold $\tau$ for the number of unique RSSI values per AP.}%
  \Output{Processed fingerprint database $\mathcal{D}$.}%
  Let $\bm{A}$ be the set of AP RSSI columns in $\mathcal{D}_{org}$.\;
  $\bm{S} \leftarrow \emptyset$\;
  \For{$a \in \bm{A}$}{%
    \If{$\mathrm{NumberOfUniqueValues}(a) > \tau$}{%
      $\bm{S} \leftarrow \bm{S} \cup a$ }%
  }%
  $\mathcal{D} \leftarrow \bm{S} \cup \left(\mathcal{D}_{org} \setminus \bm{A}\right)$\tcp*[r]{replace the RSSIs}
  \Return{$\mathcal{D}$}%
\end{algorithm}

Note that the tunable hyperparameter $\tau$ can control the trade-off between
feature-dimension reduction and information retention. As the statistical
characteristics of databases differ from one another, its value can be selected
through statistical analysis of the given database or by grid searching in
practice. For further discussions and a detailed case study of $\tau$'s impact
on indoor localization performance with the UJIIndoorLoc database, readers are
referred to our prior work~\cite{AP_SLC_01}, where a prototype of the
AP-selection algorithm is validated.

After the AP selection, we can construct the datasets defined in
\eqref{eq:datasets} from the processed fingerprint database $\mathcal{D}$.

\subsection{Pre-Training and Cloning}
\label{sec:pre-training-cloning}
As an effective initialization for the SSL training, we pre-train $m_{p}$ with
the labeled dataset $D_{L}$ over a limited number of epochs and, after the
pre-training, clone $m_{p}$ into $m_{s}$ and $m_{t}$, i.e.,
\begin{equation}
  \label{eq:pre-training-cloning}
  \bm{\theta}_{s}^{0} = \bm{\theta}_{t}^{0} = \bm{\theta}_{p},
\end{equation}
where $\bm{\theta}_{s}^{0}$ and $\bm{\theta}_{t}^{0}$ are initial vectors of
the parameters (i.e., biases and weights) of $m_{s}$ and $m_{t}$, respectively,
and $\bm{\theta}_{p}$ is a vector of the pre-trained parameters of $m_{p}$.

This process of pre-training and cloning can not only alleviate the cold start
problem~\cite{ColdStart_01,ColdStart_02} but also expedite the SSL training
process.

\subsection{SSL Training}
\label{sec:ssl-training}

\subsubsection{Batch-level Noise Injection}
\label{sec:batch-level-noise-injection}
If the fingerprint database does not provide unlabeled data, we can use the
``batch-level noise injection'' block to generate unlabeled data from the
labeled ones for SSL training. Although noise injection, often combined with
data augmentation, has been widely used in other research areas (e.g., computer
vision~\cite{aug_inject_01,aug_inject_02} and speech
recognition~\cite{aug_inject_03}) to improve the generalization capability of a
model, it has hardly been applied to indoor localization based on Wi-Fi
fingerprinting.

In generating \textit{noise-injected unlabeled RSSI data} based on the labeled
ones, we use a \textit{truncated} additive white Gaussian noise (AWGN) to
ensure the fundamental assumptions on the structure of data for
SSL~\cite{ssl:overview1}: For $i{=}1,{\ldots},n_{L}$,
\begin{equation}
  \label{eq:blni}
  \bm{x}_{N,i} = \bm{\kappa}\left(\bm{x}_{L,i} + \bm{w}, a, b\right),
\end{equation}
where
\begin{equation}
  \label{eq:awgn}
  \bm{w} \sim \mathcal{N}\left(\bm{0},\sigma^{2}\bm{I}\right),
\end{equation}
and $\bm{\kappa}(\bm{x}, a, b)$ is an element-wise truncation function whose
$j$th element $\kappa_{j}(x_{j}, a, b)$ is defined as follows: For
$j{=}1,{\ldots},n_{F}$, where $n_{F}$ is the feature dimension of the RSSI
vector $\bm{x}_{L,i}$,
\begin{equation}
  \label{eq:truncation}
  \kappa_{j}(x_{j}, a, b) =
  \begin{cases}
    a     & \text{if $x_{j} < a$,} \\
    b     & \text{if $x_{j} > b$,} \\
    x_{j} & \text{otherwise.}
  \end{cases}
\end{equation}
As the RSSI values of each batch are typically normalized to the range of $[0,
      1]$, we set $\sigma^{2}$, $a$, and $b$ to $10^{-8}$, 0, and 1, respectively, in
this paper.

\subsubsection{Training of Student and Teacher Models}
\label{sec:st-model-training}
After the pre-training and cloning, $m_{s}$ and $m_{t}$ are trained in an SSL
manner based on the Mean Teacher.

\begin{figure}[!htb]
  \centering%
  \includegraphics[angle=-90,width=.9\textwidth]{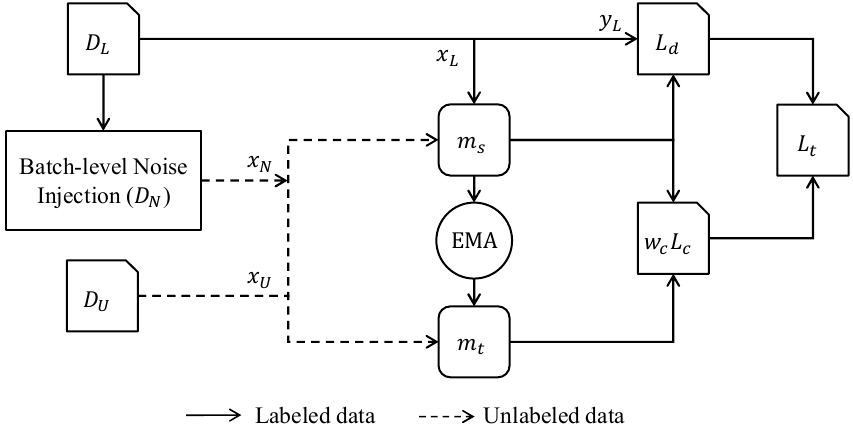}
  \caption{Details of the training of $m_{s}$ and $m_{t}$.}
  \label{fig:train}
\end{figure}

Let $\bm{f}$ be a function relating the input, the output, and the parameters
of the common DNN architecture underlying both $m_{s}$ and $m_{t}$. Then we can
define the predictions from $m_{s}$ and $m_{t}$ for the labeled and the
unlabeled datasets formulated in \eqref{eq:datasets} as follows: For $m_{s}$,
\begin{equation}
  \label{eq:student-predictions}
  \begin{aligned}
    \hat{\bm{y}}_{L,i}^{s} & = \bm{f}(\bm{x}_{L,i}, \bm{\theta}_{s}), ~~~
    i = 1, \ldots, n_{L},                                                 \\
    \hat{\bm{y}}_{U,i}^{s} & = \bm{f}(\bm{x}_{U,i}, \bm{\theta}_{s}), ~~~
    i = 1, \ldots, n_{U},                                                 \\
    \hat{\bm{y}}_{N,i}^{s} & = \bm{f}(\bm{x}_{N,i}, \bm{\theta}_{s}), ~~~
    i = 1, \ldots, n_{N},
  \end{aligned}
\end{equation}
and for $m_{t}$,
\begin{equation}
  \label{eq:teacher-predictions}
  \begin{aligned}
    \hat{\bm{y}}_{U,i}^{t} & = \bm{f}(\bm{x}_{U,i}, \bm{\theta}_{t}), ~~~
    i = 1, \ldots, n_{U},                                                 \\
    \hat{\bm{y}}_{N,i}^{t} & = \bm{f}(\bm{x}_{N,i}, \bm{\theta}_{t}), ~~~
    i = 1, \ldots, n_{N}.
  \end{aligned}
\end{equation}
Based on the predictions in \eqref{eq:student-predictions} and
\eqref{eq:teacher-predictions}, the prediction loss $L_{d}$ and the consistency
loss $L_{c}$ are defined as follows:
\begin{equation}
  \label{eq:losses}
  \begin{aligned}
    L_{d} & = \mathcal{L}_{d}\left(\hat{\bm{y}}_{L,i}^{s},\bm{y}_{L,i}\right),                                                                                       \\
    L_{c} & = \mathcal{L}_{u}\left(\hat{\bm{y}}_{U,i}^{s},\hat{\bm{y}}_{U,i}^{t}\right) + \mathcal{L}_{n}\left(\hat{\bm{y}}_{N,i}^{s},\hat{\bm{y}}_{N,i}^{t}\right),
  \end{aligned}
\end{equation}
where $\mathcal{L}_{d}$ is a loss function for the difference between the
$m_{s}$ prediction and the ground truth based on the labeled data, and
$\mathcal{L}_{u}$ and $\mathcal{L}_{n}$ are loss functions for the prediction
differences between $m_{s}$ and $m_{t}$ based on the unlabeled and the
noise-injected data, respectively. Following the Mean
Teacher~\cite{ssl:mean_teacher}, we train $m_{s}$ with the total loss $L_{t}$
based on both prediction and consistency losses, i.e.,
\begin{equation}
  L_{t} = L_{d} + w_{c}L_{c},
\end{equation}
where $w_{c}$ is a weight coefficient regulating the contribution of $L_{c}$
with respect to $L_{d}$. After the parameters of $m_{s}$ are updated by
backpropagation based on $L_{t}$, the parameters of $m_{t}$ are updated to the
EMAs of the parameters of $m_{s}$, i.e.,
\begin{equation}
  \bm{\theta}_{t}^{i}={\alpha}\bm{\theta}_{t}^{i-1}+(1-{\alpha})\bm{\theta}_{s}^{i},
\end{equation}
where $\bm{\theta}_{t}^{i}$ and $\bm{\theta}_{s}^{i}$ are the parameter vectors
of $m_{t}$ and $m_{s}$ during the $i$th ($1,2,{\ldots}$) training step,
respectively, and $\alpha~{\in}(0,1]$ is a smoothing coefficient of the EMAs.
The details of the training of the student and teacher models are shown in
Fig.~\ref{fig:train}.

After the SSL training, $m_{t}$ is used for performance evaluation with
$D_{T}$.

\subsection{Performance/Complexity Trade-off}
\label{sec:pc-trade-off}
Wi-Fi RSSI fingerprints exhibit inherent characteristics of high noise, weak
spatial correlation, and spatiotemporal fluctuation. The consistency
regularization strategy of the Mean Teacher enables the model to learn more
robust potential features of RSSI fingerprints through the interaction between
the teacher and student models, effectively suppressing noise interference. The
smooth target labels generated by the EMA of weights, too, align with the slow
spatiotemporal variation characteristics of RSSIs, enhancing the model's
adaptability to changes in wireless environments.

As the proposed SSL framework is based on the Mean Teacher, it introduces a
moderate increase in computational complexity compared to pure SL frameworks
due to its running two models during the training. Compared to existing SSL
methods, its efficiency is generally superior: For example, unlike the Temporal
Ensembling that generates targets only once per epoch, the Mean Teacher
aggregates information after each step, allowing it to scale to large datasets.
Also, though both the student and teacher models process input data in each
training step, only the student model requires backpropagation while the
parameters of the teacher model are updated via EMA, which has constant time
and space complexity of $\bigO(1)$ per parameter. As for memory footprint,
because the network architectures of both models are identical, it is roughly
double the base network size.

The AP selection described in Algorithm~\ref{alg:ap-selection} has a time
complexity of $\bigO(n_{org}^{2}\log\,n_{org})$, where $n_{org}$ is the size of
$\mathcal{D}_{org}$ (i.e., the number of RSSI fingerprints), because the time
complexity of the function \texttt{NumberOfUniqueValues} is
$\bigO(n_{org}\log\,n_{org})$~\cite{misra82:_findin}. Since it runs only once in
the beginning as shown in Fig.~\ref{fig:ts_alg}, the increase in the overall
computational complexity by the AP selection is minimal. Its impact on the SSL
training, on the other hand, is quite significant due to the reduced feature
dimensionality; for example, the batch-level noise injection, which is part of
the proposed SSL framework, benefits as its time complexity of
$\bigO(n_{F}n_{L})$ is directly proportional to the feature dimension $n_{F}$.

Note that the computational complexity of the proposed SSL framework could be
further improved by adopting techniques such as the Gaussian ramp-up strategy
(e.g.,~\cite{yang25:_unlab_insig_label_boost}), allowing the student model to
focus more on labeled data early on before investing heavily in consistency
regularization on unlabeled data, and lightweight dual-student
models~\cite{arXiv:1909.01804}, which could decrease the number of parameters
and, thereby, computational complexity by a factor of up to
42~\cite{ZHANG2025107882}.

\section{Experimental Results}
\label{sec:exp-results}
To demonstrate the feasibility of the proposed SSL framework for scalable
indoor localization under diverse application scenarios, we use both static and
dynamic fingerprint databases for the experiments, whose results are presented
in Sections~\ref{sec:exp-uji-static} and \ref{sec:exp-xjtlu-dynamic},
respectively.

\subsection{With the UJIIndoorLoc Static Database}
\label{sec:exp-uji-static}
We use the UJIIndoorLoc~\cite{Data:UJI} as a static database, which is the most
widely-used multi-building and multi-floor Wi-Fi RSSI fingerprint database and,
as such, becomes a benchmark database for multi-building and multi-floor indoor
localization in the literature. It provides 21,048 publicly-available records,
i.e., 19,937 for training and 1,111 for validation, collected at the three
multi-floor buildings of the Jaume I University in Spain, each of which
consists of RSSIs from 520 APs and 9 ground truth labels, including IDs for
building, floor, space, user, and phone, and timestamp~\cite{Data:UJI}. As the
original test dataset of the UJIIndoorLoc database has never been released to
the public, its validation dataset is used as both validation and test sets in
most studies based on the UJIIndoorLoc database~\cite{Kim:18-1}.

\begin{figure}[!htb]
  \centering%
  \includegraphics[angle=-90,width=.8\textwidth]{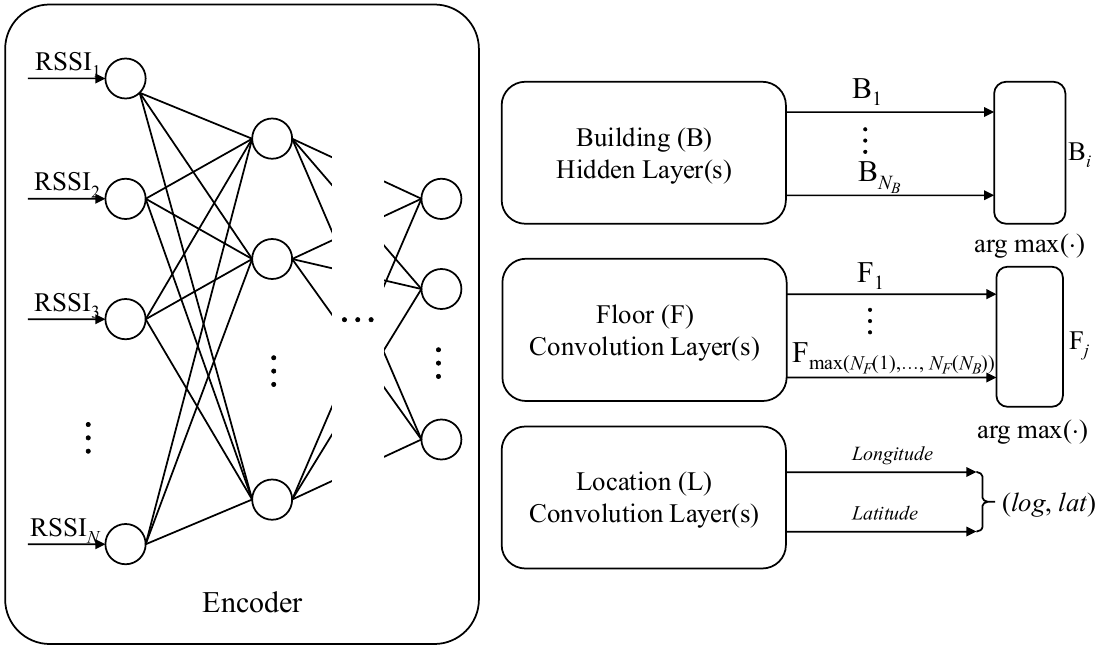}\\
  {\scriptsize (a)}\\
  \includegraphics[angle=-90,width=.8\textwidth]{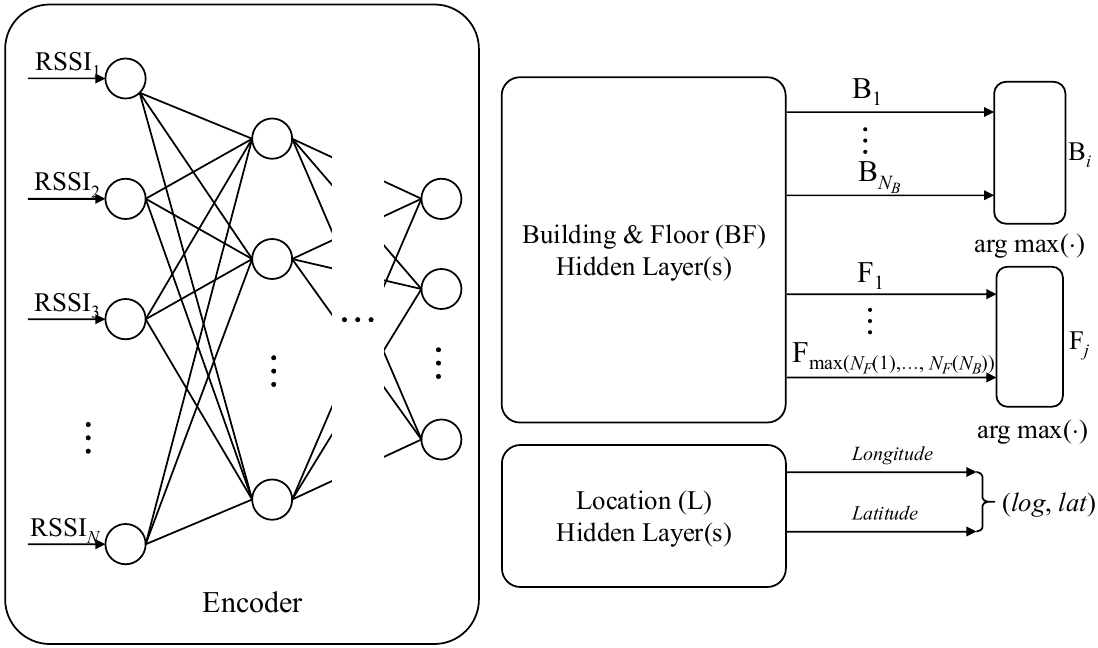}\\
  {\scriptsize (b)}\\
  \caption{Reference models based on (a) CNNLoc~\cite{cnn_01} and (b)
    SIMO-DNN~\cite{Kim:18-3}.}
  \label{fig:mdl}
\end{figure}

As reference models for the evaluation of localization performance, we use the
modified versions of SIMO-DNN~\cite{Kim:18-3} and CNNLoc~\cite{cnn_01} shown in
Figs.~\ref{fig:mdl}~(a) and (b), where, unlike the original models, an
exponential linear unit (ELU) is used as an activation function instead of a
rectified linear unit (ReLU) to avoid the vanishing gradient problem and the
dimensions of SAEs are reduced due to the use of the AP selection.

The hyperparameter values of the reference models are summarized in
Tables~\ref{tab:mdl-cnnloc} and \ref{tab:mdl-simodnn}. For training, we fix the
batch size to 16 using the Adam optimizer with a learning rate of 0.0001 for
both reference models. As for the SSL parameters, we set the values of weight
($w_{c}$) and smoothing ($\alpha$) coefficients to 1.0 and 0.999, respectively.
\begin{table}[!htb]
  \centering%
  \caption{Hyperparameter values of the CNNLoc model.}
  \label{tab:mdl-cnnloc}
  \begin{tabular}{llc}
    \toprule
                                  & \makecell[c]{Parameter}                               & Value             \\
    \midrule
    \multirow{3}{*}{Encoder}      & Hidden Layers                                         & 428,214,107       \\
                                  & Activation                                            & ELU               \\
                                  & $\mathcal{L}_{d}$                                     & MSE               \\
    \cmidrule(lr){2-3}
    \multirow{3}{*}{B Classifier} & Hidden Layers                                         & 107,107,3         \\
                                  & Activation                                            & ELU               \\
                                  & Output Activation                                     & Softmax           \\
                                  & $\mathcal{L}_{d}$                                     & CE                \\
    \cmidrule(lr){2-3}
    \multirow{3}{*}{F Classifier} & Convolution Layers                                    & 99-22,66-22,33-22 \\
                                  & Hidden Layers                                         & 2211,5            \\
                                  & Activation                                            & ELU               \\
                                  & Output Activation                                     & Softmax           \\
                                  & $\mathcal{L}_{d}$                                     & CE                \\
    \cmidrule(lr){2-3}
    \multirow{3}{*}{L Regressor}  & Convolution Layers                                    & 99-22,66-22,33-22 \\
                                  & Hidden Layers                                         & 2211,2            \\
                                  & Activation                                            & ELU               \\
                                  & Output Activation                                     & Linear            \\
                                  & $\mathcal{L}_{d}$,$\mathcal{L}_{u}$,$\mathcal{L}_{n}$ & MSE               \\
    \bottomrule
  \end{tabular}
\end{table}
\begin{table}[!htb]
  \centering%
  \caption{Hyperparameter values of the SIMO-DNN model.}
  \label{tab:mdl-simodnn}
  \begin{tabular}{llc}
    \toprule
                                   & \makecell[c]{Parameter}                               & Value         \\
    \midrule
    \multirow{3}{*}{Encoder}       & Hidden Layers                                         & 428,214,107   \\
                                   & Activation                                            & Tanh          \\
                                   & $\mathcal{L}_{d}$                                     & MSE           \\
    \cmidrule(lr){2-3}
    \multirow{3}{*}{BF Classifier} & Hidden Layers                                         & 520,520,8     \\
                                   & Activation                                            & Tanh          \\
                                   & Output Activation                                     & Sigmoid       \\
                                   & $\mathcal{L}_{d}$                                     & BCE           \\
    \cmidrule(lr){2-3}
    \multirow{3}{*}{L Regressor}   & Hidden Layers                                         & 520,520,520,2 \\
                                   & Activation                                            & Tanh          \\
                                   & Output Activation                                     & Linear        \\
                                   & $\mathcal{L}_{d}$,$\mathcal{L}_{u}$,$\mathcal{L}_{n}$ & MSE           \\
    \bottomrule
  \end{tabular}
\end{table}

We also apply the AP selection with the threshold ($\tau$) of 2 to the training
set of each case and exclude the APs not selected in the training set in the
corresponding validation and training datasets.

We evaluate the performance of multi-building and multi-floor localization
using the EvAAL 3D error~\cite{EvAAL}, which is defined as follows:
\begin{equation}
  \label{eq:EvAAL}
  \text{3D~error} = p_{\text{b}}b_{\text{miss}} + p_{\text{f}}f_{\text{miss}} + \text{2D~error}~[\si{\m}],
\end{equation}
where $p_{\text{b}}$ and $p_{\text{f}}$ are the penalties for building and
floor misclassification, which are set to \SI{50}{\m} and \SI{4}{\m},
respectively, $b_{\text{miss}}$ is an indicator variable for building
misclassification (i.e., 1 for misclassification and 0 otherwise),
$f_{\text{miss}}$ is the absolute difference between the correct and the
estimated floor IDs, and 2D error is the 2D \textit{Euclidean distance} between
the correct and the estimated locations.

To quantify the improvement of the proposed SSL framework over a conventional
SL framework in terms of the mean 3D error, we define the percentage of the
relative improvement as follows:
\begin{equation}
  \label{eq:rel-improvement}
  \eta_{\text{mean}} = \frac{E^{\text{SL}}_{\text{mean}} - E^{\text{SSL}}_{\text{mean}}}{E^{\text{SL}}_{\text{mean}}}{\times}100~[\%],
\end{equation}
where $E^{\text{SL}}_{\text{mean}}$ and $E^{\text{SSL}}_{\text{mean}}$ are the
mean 3D errors under the SL and SSL frameworks, respectively. We also define
$\eta_{\text{min}}$ and $\eta_{\text{max}}$ for the minimum and maximum 3D
errors in a similar way.

The experiments with the UJIIndoorLoc database were conducted on a workstation
with an Intel i9-13900x CPU, 64GB RAM, and NVIDIA RTX 4090 GPU running Ubuntu
20.04 LTS with Python 3.9.18 and PyTorch 2.0.

\subsubsection{Static Training Based on A Hybrid Database}
\label{sec:static-training}
Our first experimental scenario focuses on static
training~\cite{static_dynamic_training} based on a hybrid database, where a
localization model is trained based on both labeled and unlabeled fingerprint
data under the proposed SSL framework during the offline phase. In this
scenario, we assume that a fingerprint database consists of not only the
labeled data measured at pre-arranged RPs from in/outsourced collection but
also the unlabeled data measured at arbitrary locations from voluntary
contributions.

\begin{figure}[!htb]
  \centering%
  \includegraphics[angle=-90,width=.75\textwidth]{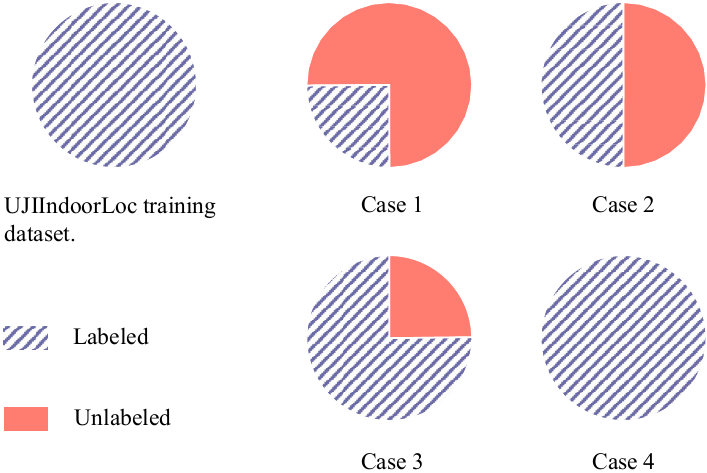}
  \caption{Preparation of four cases of labeled and unlabeled training datasets
    at RP level for static training based on a hybrid database.}
  \label{fig:exp_1}
\end{figure}

To investigate the effect of the amount of unlabeled data on localization
performance, we prepare four different cases of labeled and unlabeled training
datasets based on the training dataset of the UJIIndoorLoc database, where the
percentage of the unlabeled data, i.e., whose labels are ignored during the
training, decreases from 75 to 0 from Case~1 to 4 as shown in
Fig.~\ref{fig:exp_1}; in Case~4 where there is no unlabeled data, the proposed
framework uses the batch-level noise injection block to generate noise-injected
unlabeled data for SSL training as discussed in
Section~\ref{sec:batch-level-noise-injection}.

The localization performance of the reference models trained based on labeled
and unlabeled data under the proposed SSL framework is evaluated using the
validation dataset of the UJIIndoorLoc database. For comparison, we also train
the reference models under the SL framework using only the labeled data. The 3D
errors of the reference models under both SSL and SL frameworks and their
percentage of relative improvements are summarized in
Table~\ref{tab:res_uji_sc1}.
\begin{table*}[!ht]
  \centering%
  \caption{3D errors of the reference models under the proposed SSL and the
    conventional SL frameworks and their percentage of relative improvements for
    static training based on a hybrid database.}
  \label{tab:res_uji_sc1}
  \begin{threeparttable}
    \small
    \begin{tabular}{lllccccrr}
      \toprule
      \multirow{2}{*}{Case}            & \makecell[c]{\multirow{2}{*}{Model}} & \makecell[c]{\multirow{2}{*}{Learning}} & \multicolumn{6}{c}{3D Error [\si{\m}]}                                                                                                                          \\
      \cmidrule(lr){4-9}
                                       &                                      &                                         & Mean\tnote{*}                          & $\eta_{\text{mean}}[\%]$ & Min   & $\eta_{\text{min}}[\%]$ & \makecell[c]{Max} & \makecell[c]{$\eta_{\text{max}}[\%]$} \\
      \midrule
      \makecell[c]{\multirow{4}{*}{1}} & \multirow{2}{*}{CNNLoc}              & SL                                      & 9.698$\pm$0.11                         & \multirow{2}{*}{6.366}   & 8.892 & \multirow{2}{*}{1.889}  & 12.040            & \multirow{2}{*}{18.048}               \\
                                       &                                      & SSL                                     & 9.081$\pm$0.04                         &                          & 8.724 &                         & 9.867             &                                       \\
      \cmidrule(lr){3-9}
                                       & \multirow{2}{*}{SIMO-DNN}            & SL                                      & 9.757$\pm$0.08                         & \multirow{2}{*}{7.748}   & 9.044 & \multirow{2}{*}{6.457}  & 10.754            & \multirow{2}{*}{8.499}                \\
                                       &                                      & SSL                                     & 9.001$\pm$0.06                         &                          & 8.460 &                         & 9.840             &                                       \\
      \cmidrule(lr){2-9}
      \makecell[c]{\multirow{4}{*}{2}} & \multirow{2}{*}{CNNLoc}              & SL                                      & 9.222$\pm$0.10                         & \multirow{2}{*}{7.403}   & 8.545 & \multirow{2}{*}{3.242}  & 11.872            & \multirow{2}{*}{24.983}               \\
                                       &                                      & SSL                                     & 8.540$\pm$0.03                         &                          & 8.268 &                         & 8.906             &                                       \\
      \cmidrule(lr){3-9}
                                       & \multirow{2}{*}{SIMO-DNN}            & SL                                      & 9.052$\pm$0.04                         & \multirow{2}{*}{3.709}   & 8.627 & \multirow{2}{*}{4.080}  & 10.085            & \multirow{2}{*}{3.332}                \\
                                       &                                      & SSL                                     & 8.716$\pm$0.05                         &                          & 8.275 &                         & 9.749             &                                       \\
      \cmidrule(lr){2-9}
      \makecell[c]{\multirow{4}{*}{3}} & \multirow{2}{*}{CNNLoc}              & SL                                      & 9.016$\pm$0.10                         & \multirow{2}{*}{7.389}   & 8.409 & \multirow{2}{*}{3.520}  & 11.345            & \multirow{2}{*}{23.006}               \\
                                       &                                      & SSL                                     & 8.349$\pm$0.02                         &                          & 8.113 &                         & 8.735             &                                       \\
      \cmidrule(lr){3-9}
                                       & \multirow{2}{*}{SIMO-DNN}            & SL                                      & 8.850$\pm$0.07                         & \multirow{2}{*}{4.615}   & 8.263 & \multirow{2}{*}{3.522}  & 10.004            & \multirow{2}{*}{2.339}                \\
                                       &                                      & SSL                                     & 8.442$\pm$0.05                         &                          & 7.972 &                         & 9.770             &                                       \\
      \cmidrule(lr){2-9}
      \makecell[c]{\multirow{4}{*}{4}} & \multirow{2}{*}{CNNLoc}              & SL                                      & 8.935$\pm$0.09                         & \multirow{2}{*}{6.223}   & 8.325 & \multirow{2}{*}{3.387}  & 10.559            & \multirow{2}{*}{7.520}                \\
                                       &                                      & SSL                                     & 8.379$\pm$0.05                         &                          & 8.043 &                         & 9.765             &                                       \\
      \cmidrule(lr){3-9}
                                       & \multirow{2}{*}{SIMO-DNN}            & SL                                      & 8.843$\pm$0.07                         & \multirow{2}{*}{5.256}   & 8.238 & \multirow{2}{*}{4.188}  & 10.276            & \multirow{2}{*}{9.508}                \\
                                       &                                      & SSL                                     & 8.379$\pm$0.06                         &                          & 7.893 &                         & 9.299             &                                       \\
      \bottomrule
    \end{tabular}
    \begin{tablenotes}
      \item[*] Mean error with 95\% confidence interval.
    \end{tablenotes}
  \end{threeparttable}
\end{table*}

As shown in Table~\ref{tab:res_uji_sc1}, the proposed SSL framework
consistently outperforms the conventional SL framework for all the cases,
demonstrating the effectiveness of the proposed SSL framework for scalable
indoor localization. Specifically, the percentage of relative improvement in
mean 3D error (i.e., $\eta_{\text{mean}}$) ranges from 3.709\% to 7.748\% for
the SIMO-DNN and from 6.223\% to 7.403\% for the CNNLoc models, respectively,
which indicates the effectiveness of the proposed SSL framework in enhancing
the localization performance of the reference models even in Case~4 where there
are no unlabeled data. Moreover, the proposed SSL framework shows a significant
improvement in the maximum 3D error, especially for the CNNLoc models, which is
more crucial to real-world applications than the mean 3D error.

\begin{figure}[!htb]
  \centering%
  \includegraphics[width=.95\textwidth,trim={0 100 0 90},clip]{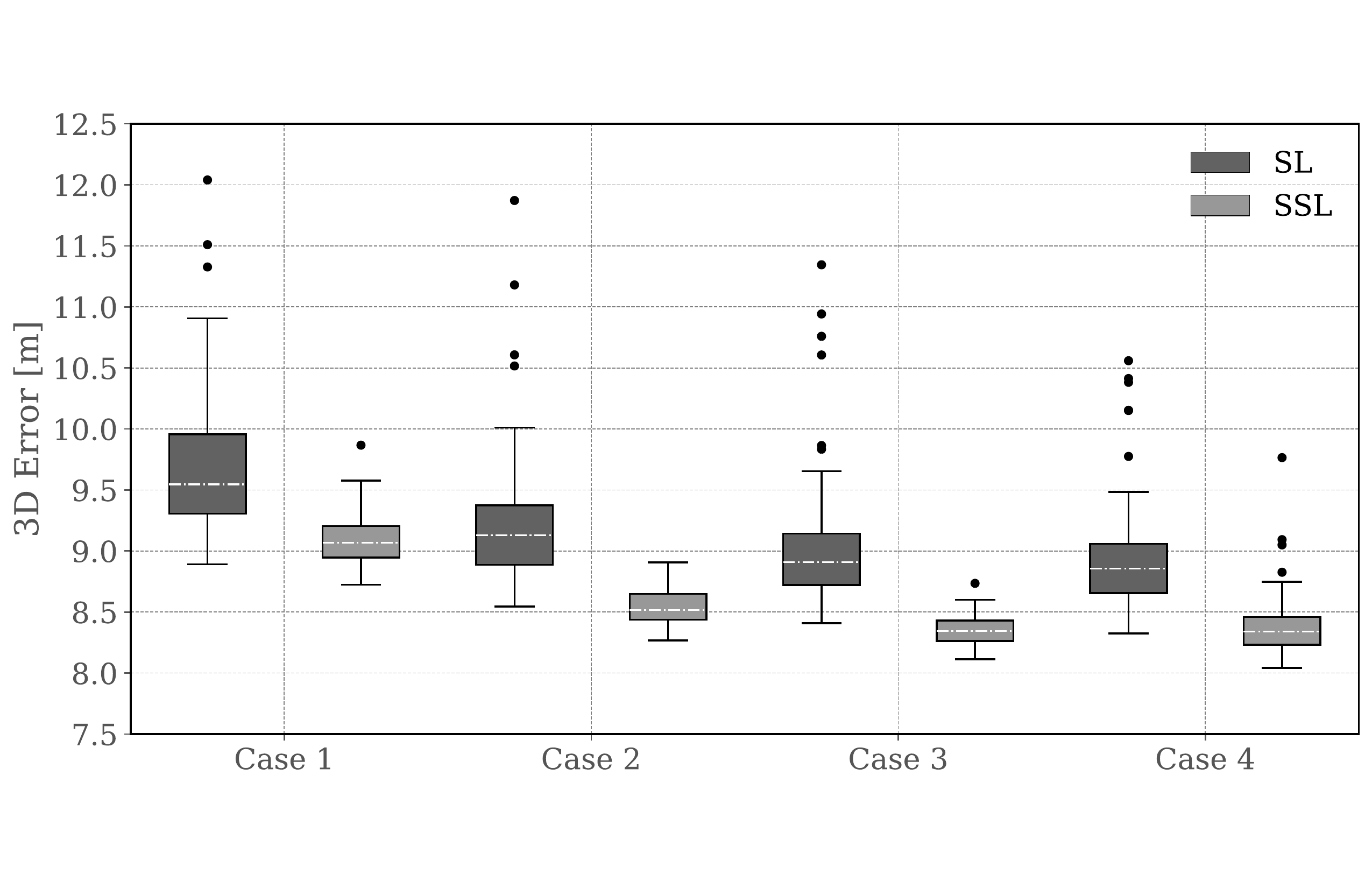}\\
  {\scriptsize (a)}\\
  \includegraphics[width=.95\textwidth,trim={0 100 0 90},clip]{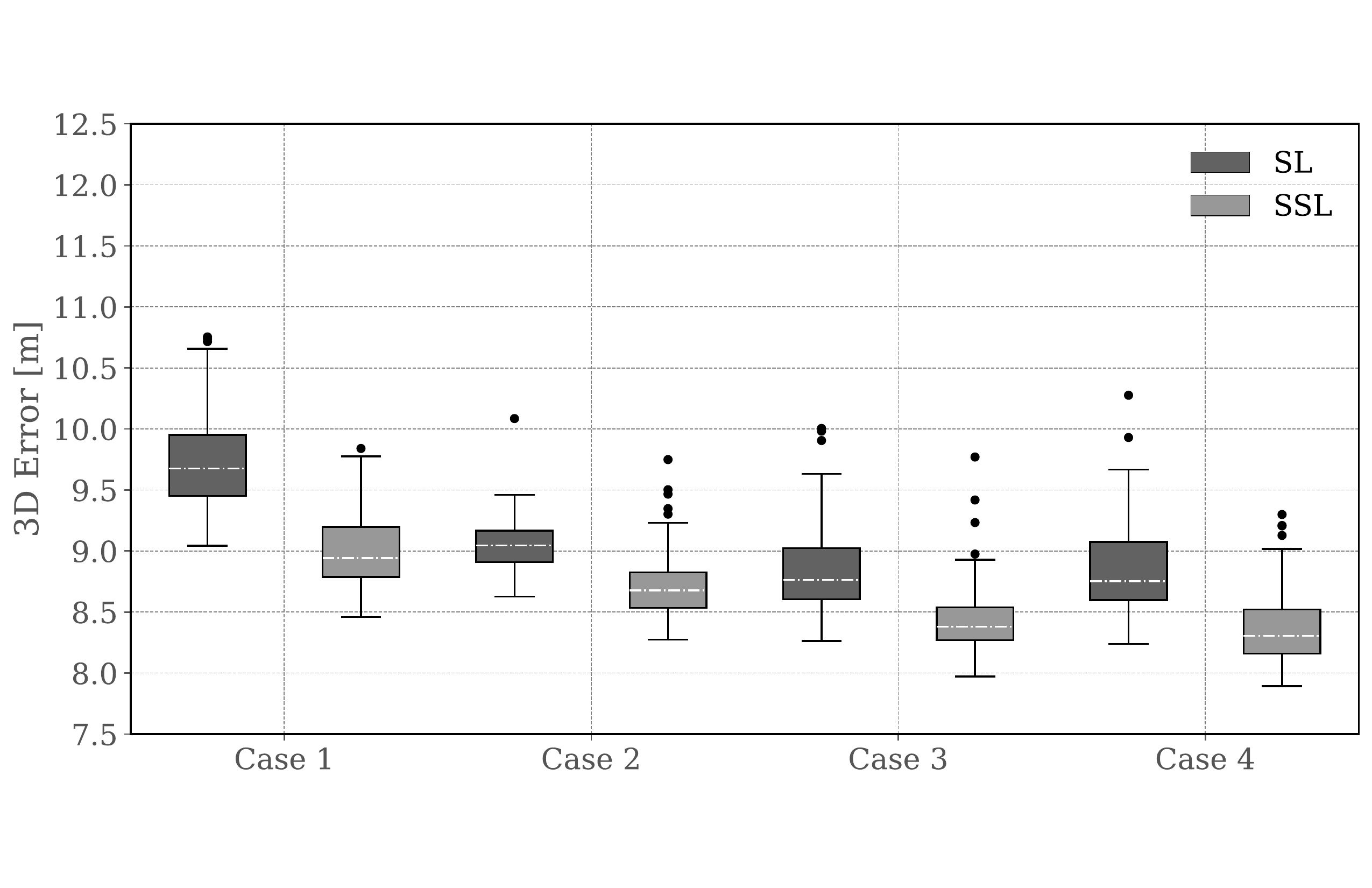}\\
  {\scriptsize (b)}\\
  \caption{Box plots of the 3D errors of the reference models under the proposed
    SSL and the conventional SL frameworks for static training based on a hybrid
    database: (a) CNNLoc and (b) SIMO-DNN.}
  \label{fig:res_1}
\end{figure}

The box plots of the 3D errors in Fig.~\ref{fig:res_1} better visualize the
improvement of the indoor localization performance by the proposed SSL
framework.

In comparison to the conventional SL framework, the proposed SSL framework not
only provides lower mean, minimum, and maximum 3D errors but also reduces
interquartile ranges (IQRs) of 3D errors over all the cases, which are a
measure of data spread defined as the difference between the 75th and the 25th
quartiles and, as such, demonstrate the robustness of the location estimations
by the proposed SSL framework in static training based on a hybrid database. As
shown in Fig.~\ref{fig:res_1}~(a) and (b), the reduction in IQRs is more
significant for the CNNLoc model than the SIMO-DNN model. Also, the comparison
of the results for Case~2 and Case~4 shows that the 3D errors of the reference
models trained with only 50\% of the labeled data under the proposed SSL
framework can achieve better performance than those trained with the entirety
of the labeled data under the conventional SL framework, which implies that the
proposed SSL framework could significantly reduce the labor cost for the
construction of fingerprint databases by efficiently leveraging unlabeled data
in training localization models.

In Table~\ref{tab:comp-sota}, we compare the multi-building and multi-floor
indoor localization performance of the proposed models for Case~4 (i.e., with
100\% of the labeled data) with that of state-of-the-art localization models
using the UJIIndoorLoc database.
\begin{table}[!htb]
  \centering%
  \caption{3D errors of the reference and the state-of-the-art indoor localization
    models based on the UJIIndoorLoc database.}
  \label{tab:comp-sota}
  \begin{threeparttable}
    \begin{tabular}{lc}
      \toprule
      \makecell[c]{Model}                                                        & 3D error [\si{\m}]\tnote{*} \\
      \midrule
      MOSAIC Fixed~\cite{wknn-02}\tnote{\dag}                                    & 9.01                        \\
      HFTS~\cite{EvAAL}\tnote{\dag}                                              & 8.49                        \\
      ICSL~\cite{EvAAL}\tnote{\dag}                                              & 7.67                        \\
      RTLS@UM~\cite{EvAAL}\tnote{\dag}                                           & 6.20                        \\
      \midrule
      Scalable DNN~\cite{Kim:18-1}                                               & 9.29                        \\
      CNNLoc~\cite{cnn_01}\tnote{\ddag}                                          & 11.78                       \\
      K-Means wKNN~\cite{wknn-KMeans}                                            & 8.62                        \\
      Hierarchical RNN~\cite{rela_rnn_01}                                        & 8.62                        \\
      MOGP RNN~\cite{rela_rnn_02}\tnote{\ddag}                                   & 8.42                        \\
      DumbLoc~\cite{DumbLoc}                                                     & 8.45                        \\
      \midrule
      W-GAN~\cite{rela_WGAN}\tnote{\ddag} \tnote{**}                             & 4.07                        \\
      Selective GAN-based Data Augmented DNN~\cite{wGan}\tnote{\ddag} \tnote{**} & 3.91                        \\
      SE-Loc~\cite{SE-Loc}\tnote{**}                                             & 7.29                        \\
      \midrule
      \textbf{Case 4 - CNNLoc SL}                                                & \textbf{8.33}               \\
      \textbf{Case 4 - SIMO-DNN SL}                                              & \textbf{8.24}               \\
      \textbf{Case 4 - CNNLoc SSL}                                               & \textbf{8.04}               \\
      \textbf{Case 4 - SIMO-DNN SSL}                                             & \textbf{7.89}               \\
      \bottomrule
    \end{tabular}
    \begin{tablenotes}
      \item[*] Based on the best run for fair comparison with the results based on
      a single run in the literature.
      \item [**] Only dedicated selected floors are considered for testing, which is
      not a multi-building and multi-floor localization.
      \item[\dag] Based on the testing dataset of the UJIIndoorLoc database provided
      only to the participants of the 2015 EvAAL competition~\cite{EvAAL}.
      \item[\ddag] Based on RSSI data augmentation.
    \end{tablenotes}
  \end{threeparttable}
\end{table}

Note that, as the top four models are tested based on the original testing
dataset of the UJIIndoorLoc database that is not released to the public, a fair
comparison with the rest of the models is not possible.

We observe that, among the models trained and tested based on the
publicly-available datasets of the UJIIndoorLoc database, the reference DNN
models provide better performance than the existing ones. In particular, the
SIMO-DNN model under the proposed SSL framework achieves the 3D error of
\SI{7.89}{\m}, which, to the best of the authors' knowledge, is the minimum 3D
error ever obtained by DNN-based models using the publicly-available datasets
of the UJIIndoorLoc database. In fact, Tables~\ref{tab:res_uji_sc1} and
\ref{tab:comp-sota} show that not only the minimum but also the mean 3D errors
of both reference models under the proposed SSL framework are lower than the
best results from the state-of-the-art models, including those based on RSSI
data augmentation. This demonstrates the robustness and reliability of the
proposed SSL framework for indoor localization services.

\subsubsection{Dynamic Training During the Online Phase}
\label{sec:dynamic-training}
The proposed SSL framework also enables dynamic
training~\cite{static_dynamic_training} during the online phase, where the
localization model is continuously retrained based on the unlabeled data from
users of an indoor localization system deployed in the field during the online
phase for the enhancement of its long-term performance. Therefore, in our
second scenario, we evaluate the performance improvement due to dynamic
training during the online phase enabled by the proposed SSL framework.

\begin{figure}[!htb]
  \begin{center}
    \includegraphics[angle=-90,width=.9\textwidth]{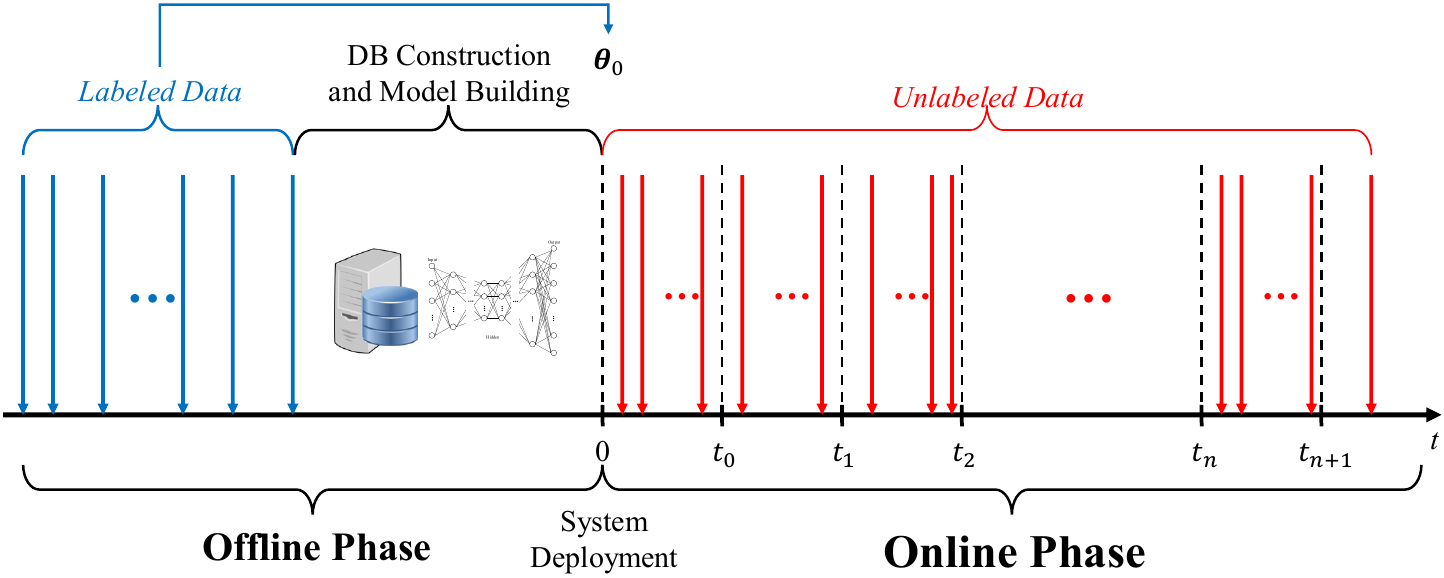}\\
    {\scriptsize (a)}\\
    \includegraphics[angle=-90,width=.9\textwidth]{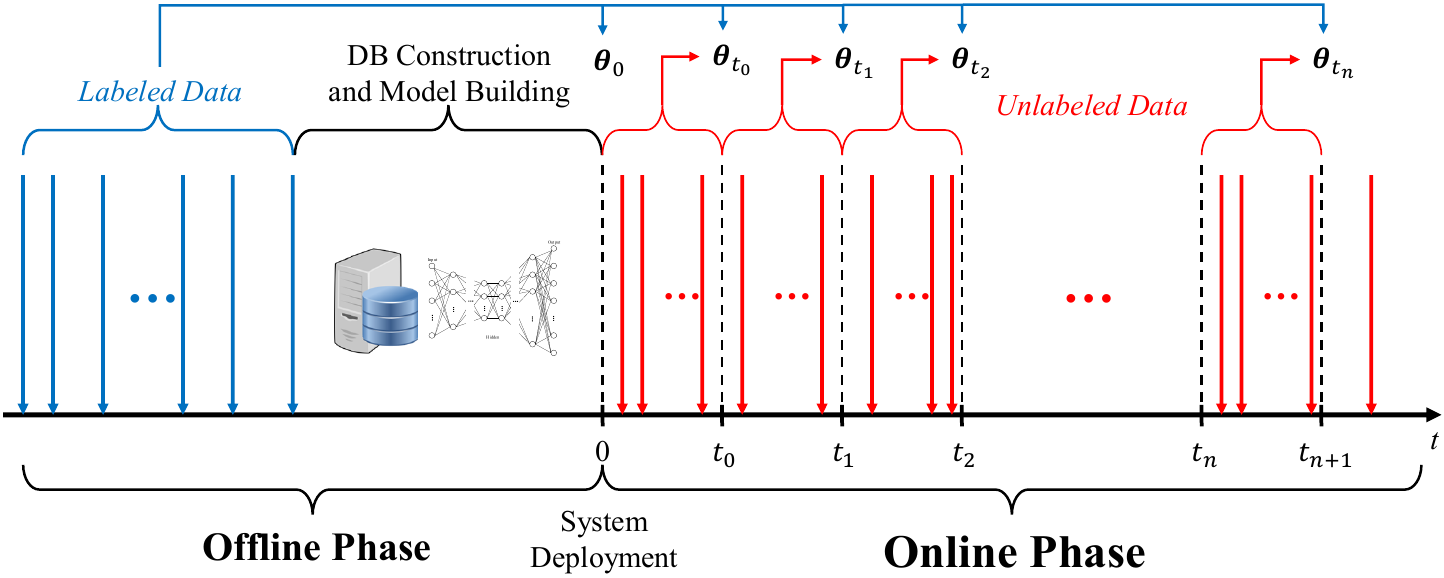}\\
    {\scriptsize (b)}\\
  \end{center}
  \caption{Training of a model under (a) the conventional SL and (b) the
    proposed SSL frameworks during the offline and the online phases, where
    $\bm{\theta}_{t}$ denotes the weights of a model trained at time $t$.}
  \label{fig:ts_comp}
\end{figure}

As shown in Fig.~\ref{fig:ts_comp}~(a), once trained during the offline phase,
the weights of the localization model are fixed throughout the online period
(i.e., $\bm{\theta}_{t_{0}}$) under the SL framework. Under the proposed SSL
framework shown in Fig.~\ref{fig:ts_comp}~(b), on the other hand, the weights
of the localization model are updated continuously based on the newly-received
unlabeled RSSIs during the online phase (i.e.,
$\bm{\theta}_{t_{i}},~i{=}1,2,{\ldots}$) to continuously reflect the
time-varying RSSI characteristics and thereby sustain the performance of the
localization model throughout the entire service period.

\begin{figure}[!htb]
  \centering%
  \includegraphics[angle=-90,width=.75\textwidth]{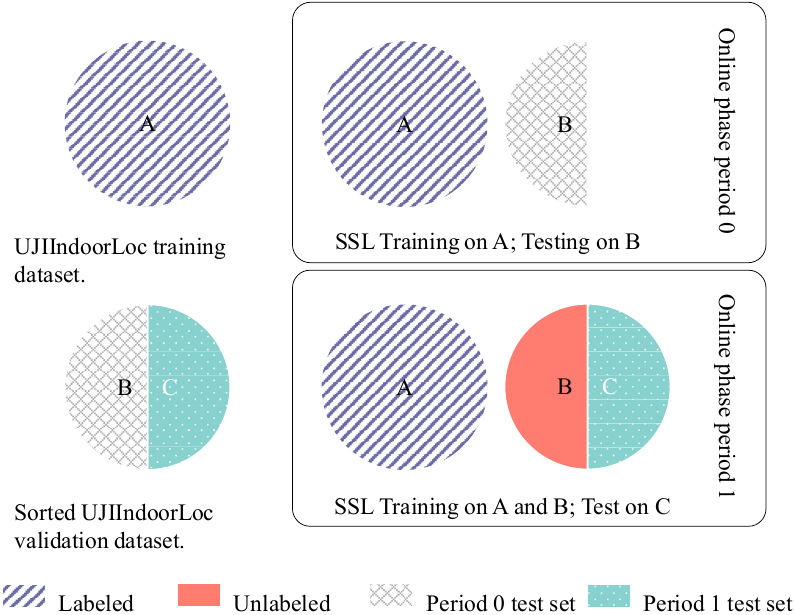}
  \caption{Preparation of the online and offline datasets for dynamic training
    during the online phase. The legends of ``labeled'' and ``unlabeled''
    indicate their use in training the model, and a nine-day temporal gap
    separates the dataset~B for Period~0 from the dataset~C for Period~1.}
  \label{fig:exp_2}
\end{figure}

For this scenario, we split the validation dataset of the UJIIndoorLoc database
into two based on the sorted timestamps of the records as shown in
Fig.~\ref{fig:exp_2}; the earlier (i.e., dataset~B for Period~0) and the later
(i.e., dataset~C for Period~1) records, which are separated by a temporal gap of
nine days, are used as the online unlabeled data from the users for SSL training
and the testing dataset for performance evaluation, respectively.

For Period~0, the reference models trained on the labeled dataset~A during the
offline phase are tested on the dataset~B under both SL and SSL frameworks. For
Period~1, however, the reference models are retrained with the unlabeled
dataset~B under the SSL framework and tested on the dataset~C, while the same
reference models trained on the labeled dataset~A during the offline phase are
tested on the dataset~C without retraining under the SL framework.

The 3D errors of the reference models under both SSL and SL frameworks and
their percentage of relative improvements are summarized in
Table~\ref{tab:res_2}.
\begin{table*}[!ht]
  \centering%
  \caption{3D errors of the reference models under the proposed SSL and the
    conventional SL frameworks and their percentage of relative improvements for
    dynamic training during the online phase.}
  \label{tab:res_2}
  \begin{threeparttable}
    \small%
    \begin{tabular}{lllcccccc}
      \toprule
      \multirow{2}{*}{Period}          & \multicolumn{1}{c}{\multirow{2}{*}{Model}} & \makecell[c]{\multirow{2}{*}{Strategy}} & \multicolumn{6}{c}{3D Error [\si{\m}]}                                                                                                                           \\
      \cmidrule(lr){4-9}
                                       &                                            &                                         & Mean\tnote{*}                          & $\eta_{\text{mean}}[\%]$ & Min    & $\eta_{\text{min}}[\%]$ & \makecell[c]{Max} & \makecell[c]{$\eta_{\text{max}}[\%]$} \\
      \midrule
      \makecell[c]{\multirow{4}{*}{0}} & \multirow{2}{*}{CNNLoc}                    & SL                                      & 7.256$\pm$0.11                         & \multirow{2}{*}{5.940}   & 6.781  & \multirow{2}{*}{5.265}  & 8.237             & \multirow{2}{*}{11.521}               \\
                                       &                                            & SSL                                     & 6.825$\pm$0.06                         &                          & 6.424  &                         & 7.288             &                                       \\
      \cmidrule(lr){3-9}
                                       & \multirow{2}{*}{SIMO-DNN}                  & SL                                      & 7.425$\pm$0.09                         & \multirow{2}{*}{7.273}   & 6.949  & \multirow{2}{*}{6.303}  & 8.036             & \multirow{2}{*}{4.704}                \\
                                       &                                            & SSL                                     & 6.885$\pm$0.08                         &                          & 6.511  &                         & 7.658             &                                       \\
      \cmidrule(lr){2-9}
      \makecell[c]{\multirow{4}{*}{1}} & \multirow{2}{*}{CNNLoc}                    & SL                                      & 10.488$\pm$0.13                        & \multirow{2}{*}{4.310}   & 9.853  & \multirow{2}{*}{1.472}  & 11.474            & \multirow{2}{*}{8.532}                \\
                                       &                                            & SSL                                     & 10.036$\pm$0.07                        &                          & 9.708  &                         & 10.495            &                                       \\
      \cmidrule(lr){3-9}
                                       & \multirow{2}{*}{SIMO-DNN}                  & SL                                      & 10.487$\pm$0.09                        & \multirow{2}{*}{7.238}   & 10.045 & \multirow{2}{*}{7.048}  & 11.417            & \multirow{2}{*}{12.297}               \\
                                       &                                            & SSL                                     & 9.728$\pm$0.06                         &                          & 9.337  &                         & 10.013            &                                       \\
      \bottomrule
    \end{tabular}
    \begin{tablenotes}
      \item[*] Mean error with 95\% confidence interval.
    \end{tablenotes}
  \end{threeparttable}
\end{table*}

As shown in Table~\ref{tab:res_2}, the proposed SSL framework consistently
surpasses the conventional SL framework in both periods, again highlighting the
effectiveness of the proposed SSL framework for dynamic training during the
online phase. As in the first scenario, the proposed SSL framework can reduce
the maximum 3D error up to 11.521\% in Period~0 and 12.297\% in Period~1,
respectively, which is more crucial to real-world applications than the mean 3D
error.

\begin{figure}[!htb]
  \centering%
  \includegraphics[width=.95\textwidth,trim={0 100 0 90},clip]{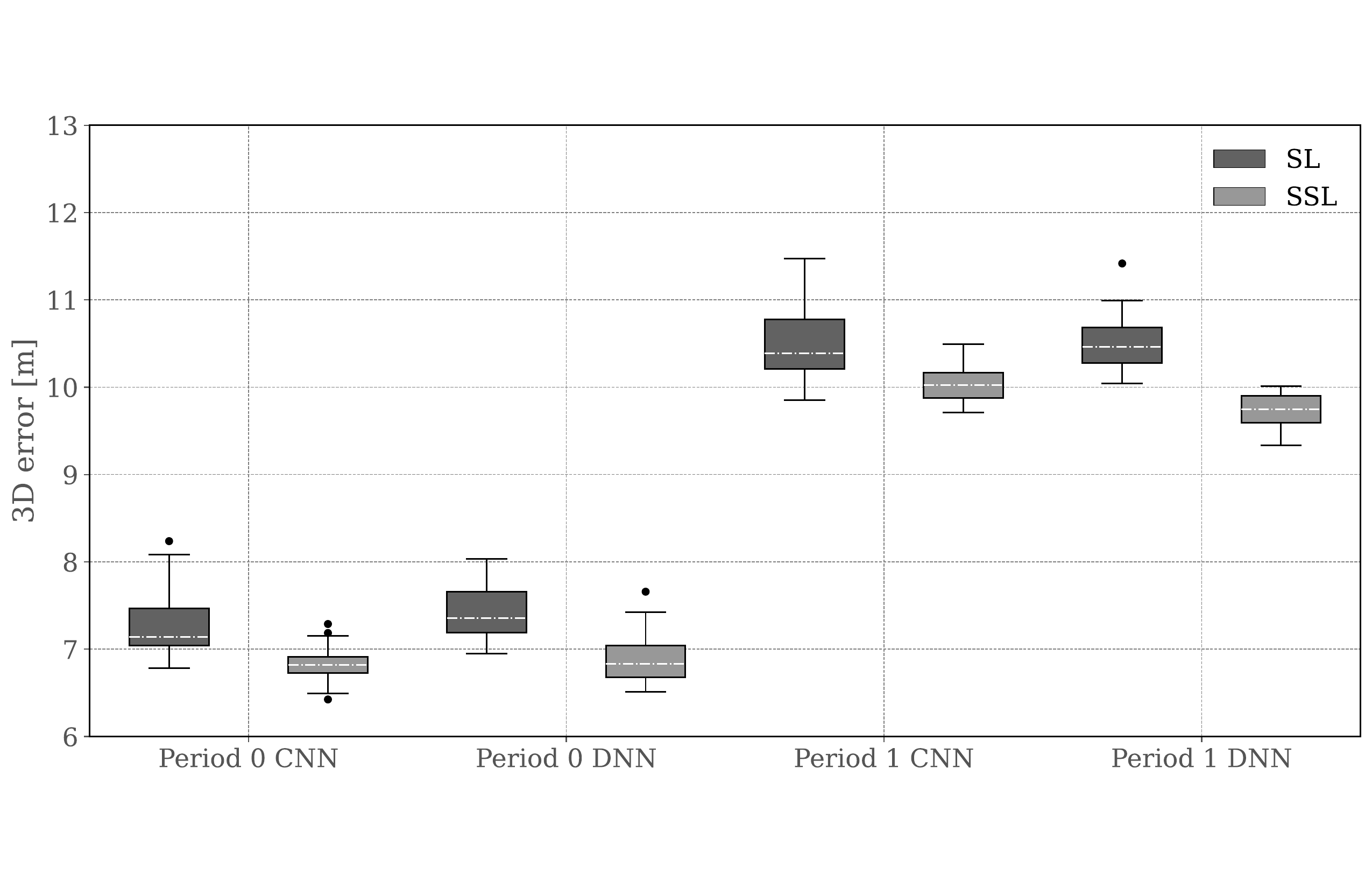}
  \caption{Box plots of the 3D errors of the reference models under the proposed
    SSL and the conventional SL frameworks for dynamic training during the
    online phase.}
  \label{fig:res_2}
\end{figure}

Like Fig.~\ref{fig:res_1}, the box plots of the 3D errors in
Fig.~\ref{fig:res_2} better visualize the improvement of the indoor
localization performance by the proposed SSL framework.

Regarding the robustness of the proposed SSL framework in dynamic training, the
IQR depicted in Fig.~\ref{fig:res_2} demonstrates a significant reduction
compared to the conventional SL framework.

\begin{figure}[!htb]
  \centering%
  \includegraphics[width=.55\textwidth,trim=80 20 35 85,clip]{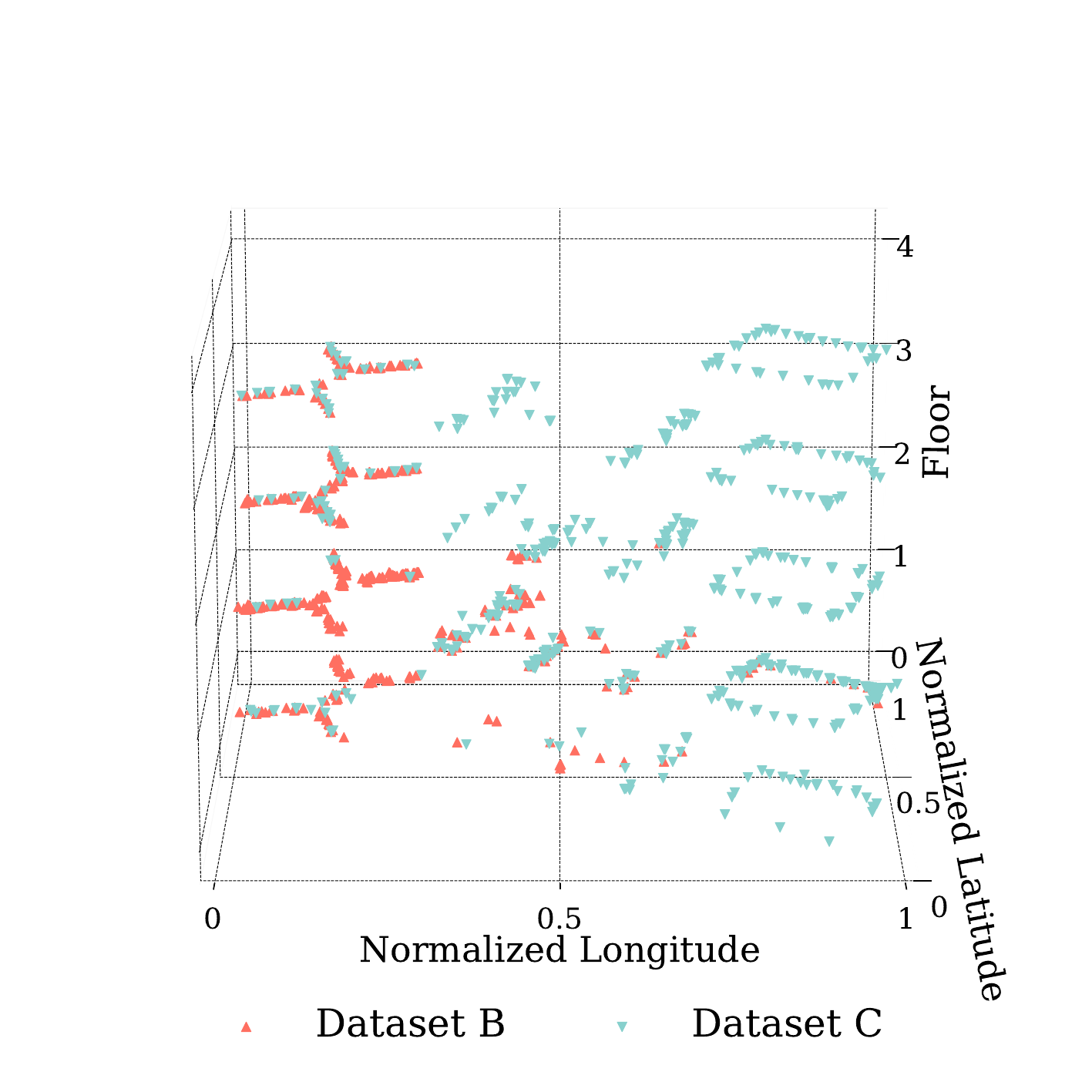}
  \caption{Spatial distributions of the RSSIs of the datasets~B and C for
    dynamic training during the online phase.}
  \label{fig:dist_exp_2}
\end{figure}

Note that the 3D errors for Period~1 significantly increase compared with those
for Period~0 under both the SL and SSL frameworks. As discussed in detail
in~\cite{AP_SLC_01}, the statistical characteristics of the RSSIs in the
training dataset (i.e., the dataset~A) and those in the validation dataset
(i.e., the datasets~B and C) of the UJIIndoorLoc database show substantial
differences between the two. Although they are from the original validation
dataset of the UJIIndoorLoc database, the two datasets B and C also show quite
different spatial coverage as shown in Fig.~\ref{fig:dist_exp_2}.

That is why the SSL framework may not significantly improve the performance of
the reference models in spite of the inclusion of dataset~B in the dynamic
training during Period~1. Such differences in the spatial distributions of the
datasets could happen in real-world applications, where new data are submitted
by users from the areas not already covered in prior datasets. In this regard,
the use of the UJIIndoorLoc database for more realistic usage scenarios like
this dynamic training during the online phase is problematic, which calls for
new types of RSSI fingerprint databases as we discuss in
Section~\ref{sec:exp-xjtlu-dynamic}.

Even though there are substantial differences in the statistical
characteristics of the RSSIs in the datasets, the overall results, under the
second scenario, demonstrate that the proposed SSL framework can effectively
enhance the performance of the indoor localization model through dynamic
training during the online phase.

\subsubsection{Limitations of the UJIIndoorLoc Database}
\label{sec:limit-ujiind-datab}
Although the UJIIndoorLoc database is a classic benchmark database for
multi-building and multi-floor localization, temporal variations in the wireless
environment were not seriously considered during its collection, and there exist
significant spatial coverage differences between Datasets B and C as shown in
Fig.~\ref{fig:dist_exp_2}, resulting in modest performance improvement in
dynamic training scenarios; this further confirms the necessity of constructing
the XJTLU dynamic database used in this study, and the experimental results
based on the XJTLU database more accurately reflect the effectiveness of the
proposed SSL framework in real dynamic wireless environments.

\subsection{With the XJTLU Dynamic Database}
\label{sec:exp-xjtlu-dynamic}
\begin{figure*}[!htb]
  \centering%
  \includegraphics[width=\textwidth]{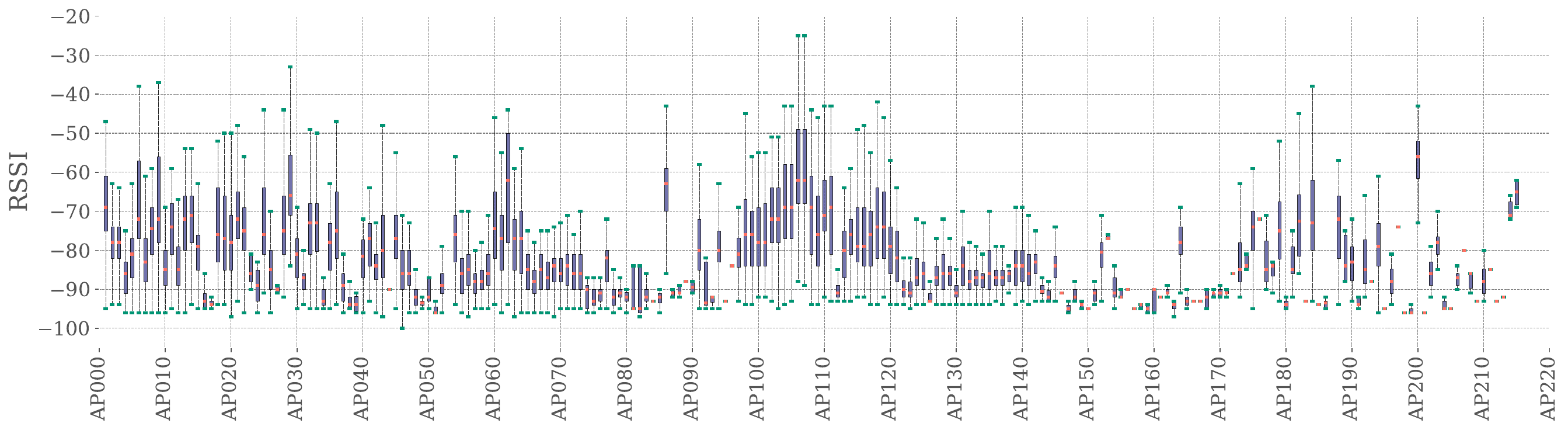}
  \caption{Box plots of the RSSIs of the training dataset of the XJTLU dynamic
    database.}
  \label{fig:exp_3}
\end{figure*}

As pointed out in Section~\ref{sec:limit-ujiind-datab}, the UJIIndoorLoc
database is not fully suitable for dynamic training scenarios. To better
investigate the effectiveness of the proposed SSL framework for dynamic as well
as static training scenarios, we present preliminary experimental results using
a subset of the XJTLU dynamic database~\cite{dynamic_static}, which is currently
under construction. This subset covers the sixth floor of the International
Research Center (IR) building on the south campus of Xi'an Jiaotong-Liverpool
University, comprising the RSSIs from 219 APs measured at 28 RPs over 44
continuous days. Fig.~\ref{fig:exp_3} summarizes the RSSI distributions of the
training dataset using box plots, excluding artificial RSSI values for
undetected APs (e.g., ``100'' in the UJIIndoorLoc database).
As indicated by their narrower ranges of the box plots in the figure, several
APs provide only a few unique RSSI values; given that the total number of APs is
limited and the total number of samples is small, we set the value of threshold
($\tau$) to 0 for the AP selection in the experiments based on the XJTLU dynamic
database so that only the APs not detected in the training dataset are excluded
(e.g., some new APs may be detected in the periods after the training dataset is
collected) through the AP selection described in
Algorithm~\ref{alg:ap-selection}, which could provide better feature alignment
during each period as well as dimensionality reduction and realistic performance
evaluation in the experiments based on the XJTLU dynamic database.

\begin{figure}[!htb]
  \centering%
  \includegraphics[angle=-90,width=.65\textwidth]{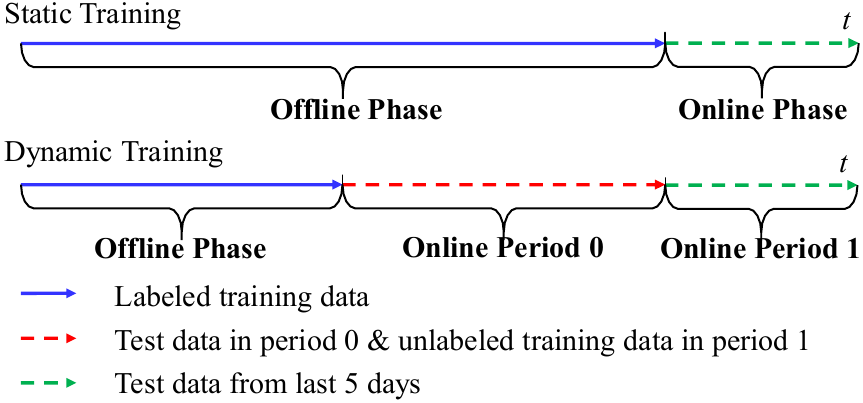}
  \caption{Experimental setup for static and dynamic training scenarios based on
    the XJTLU dynamic database.}
  \label{fig:exp_5}
\end{figure}

As for the datasets, we reserve the data measured during the last five days as
a testing dataset for performance evaluation and use the rest as training and
validation datasets. For a static training scenario, we specifically focus on
the ``Case 4'' of the cases in Section~\ref{sec:static-training}, where all
data from the database are labeled and unlabeled ones are generated by the
batch-level noise injection block for the SSL training; for this scenario, we
split the remaining data measured during the first thirty nine days based on
their timestamps and take the earlier 95\% of them as a training dataset and
the latter 5\% as a validation dataset. For a dynamic training scenario, we
split the same data into three parts again based on their timestamps: the
earlier 45\% and the next 5\% become labeled training and validation datasets
during the offline phase, and the remaining 50\% serves as an unlabeled dataset
for Period~0 during the online phase. Fig.~\ref{fig:exp_5} illustrates the
details of the dataset preparation for the two scenarios based on the XJTLU
dynamic database.

As a reference model for the evaluation of single-floor localization, we use
the modified version of the simple DNN model~\cite{dynamic_static} composed of
an encoder for noise and dimensionality reduction and a regressor for 2D
location coordinates, whose hyperparameter values are summarized in
Table~\ref{tab:para_xjt}.
\begin{table}[!htb]
  \centering%
  \caption{Hyperparameter values of the simple DNN model.}
  \label{tab:para_xjt}
  \begin{tabular}{clc}
    \toprule
    \makecell[c]{Block name}   & \makecell[c]{Parameter}                               & \makecell[c]{Value} \\
    \midrule
    \multirow{3}{*}{Encoder}   & Encoder layers                                        & 213,106,71          \\
                               & Activation                                            & ELU                 \\
                               & $\mathcal{L}_{d}$                                     & MSE                 \\
    \cmidrule(lr){2-3}
    \multirow{4}{*}{Regressor} & Hidden layers                                         & 128,128,128         \\
                               & Output layer                                          & 2                   \\
                               & Activation                                            & ELU                 \\
                               & $\mathcal{L}_{d}$,$\mathcal{L}_{u}$,$\mathcal{L}_{n}$ & MSE                 \\
    \bottomrule
  \end{tabular}
\end{table}

The model uses Adam as the optimizer with the learning rate of 0.001 and the
batch size of 8. As for the SSL parameters, we set the values of weight
($w_{c}$) and smoothing ($\alpha$) coefficients to 1.0 and 0.999, respectively,
as we did for the experiments based on the UJIIndoorLoc database. Regarding the
AP selection, on the other hand, we set the value of threshold ($\tau$) to 0 so
that the APs not detected in the training set are excluded in the corresponding
validation and training datasets.

The experiments with the XJTLU dynamic database were conducted on another
workstation equipped with an Intel i7-13700K CPU, 32GB RAM, and NVIDIA RTX 4070
GPU running Windows 11 using PyTorch 2.0 and Python 3.9.18 for cross-platform
verification. The 2D errors of the simple DNN model under both SSL and SL
frameworks and their percentage of relative improvements are summarized in
Table~\ref{tab:res_4}.
\begin{table*}[!ht]
  \centering%
  \small%
  \caption{2D errors of the simple DNN model under the proposed SSL and the
    conventional SL frameworks and their percentage of relative improvements
    using the XJTLU dynamic database.}
  \label{tab:res_4}
  \begin{threeparttable}
    \begin{tabular}{llccccrr}
      \toprule
      \multirow{2}{*}{Case/Period}            & \makecell[c]{\multirow{2}{*}{Strategy}} & \multicolumn{6}{c}{2D Error [\si{\m}]}                                                                                                                          \\
      \cmidrule(lr){3-8}
                                              &                                         & Mean\tnote{*}                          & $\eta_{\text{mean}}[\%]$ & Min   & $\eta_{\text{min}}[\%]$ & \makecell[c]{Max} & \makecell[c]{$\eta_{\text{max}}[\%]$} \\
      \midrule
      \makecell[c]{\multirow{2}{*}{Case 4}}   & SL                                      & 2.311$\pm$0.03                         & \multirow{2}{*}{17.303}  & 1.938 & \multirow{2}{*}{10.372} & 2.430             & \multirow{2}{*}{7.737}                \\
                                              & SSL                                     & 1.911$\pm$0.04                         &                          & 1.737 &                         & 2.242             &                                       \\
      \cmidrule(lr){2-8}
      \makecell[c]{\multirow{2}{*}{Period~0}} & SL                                      & 3.227$\pm$0.26                         & \multirow{2}{*}{22.344}  & 2.399 & \multirow{2}{*}{11.630} & 5.822             & \multirow{2}{*}{49.227}               \\
                                              & SSL                                     & 2.506$\pm$0.07                         &                          & 2.120 &                         & 2.956             &                                       \\
      \cmidrule(lr){2-8}
      \makecell[c]{\multirow{2}{*}{Period~1}} & SL                                      & 3.375$\pm$0.22                         & \multirow{2}{*}{17.369}  & 2.639 & \multirow{2}{*}{4.130}  & 5.383             & \multirow{2}{*}{39.755}               \\
                                              & SSL                                     & 2.789$\pm$0.04                         &                          & 2.530 &                         & 3.243             &                                       \\
      \bottomrule
    \end{tabular}
    \begin{tablenotes}
      \item[*] Mean error with 95\% confidence interval.
    \end{tablenotes}
  \end{threeparttable}
\end{table*}

As shown in Table~\ref{tab:res_4}, the proposed SSL framework outperforms the
conventional SL framework in all the cases. Specifically, the percentage of
relative improvement in maximum 2D error (i.e., $\eta_{max}$) is 49.227\% for
Period~0 and 39.755\% for Period~1 under the dynamic training scenario. The
results based on the XJTLU dynamic database again confirm the effectiveness of
the proposed SSL framework in enhancing indoor localization performance,
especially in the dynamic training scenario where the wireless environment is
subject to changes or when the model needs to be continuously updated with
newly-submitted unlabeled data during the online phase.

\begin{figure}[!htb]
  \centering%
  \includegraphics[width=.95\textwidth,trim={0 100 0 90},clip]{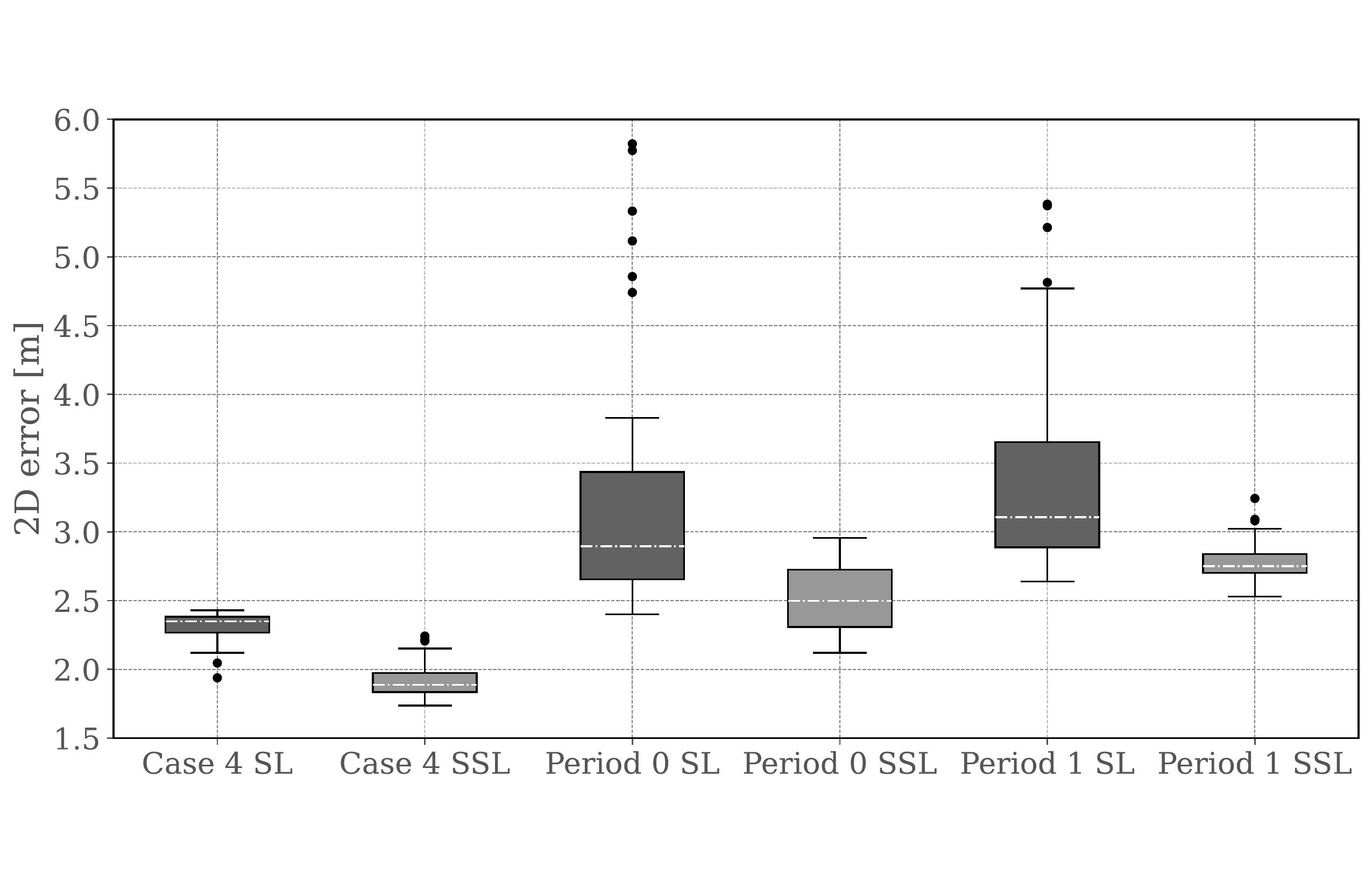}
  \caption{Box plots of the 2D errors of the simple DNN models under the
    proposed SSL and the conventional SL frameworks for static and dynamic
    training scenarios.}
  \label{fig:res_3}
\end{figure}

Fig.~\ref{fig:res_3} shows the box plots of the 2D error of the simple DNN for
each case to better visualize the improvement of the indoor localization
performance by the proposed SSL framework.

It is clear from the results that the simple DNN model under the proposed SSL
framework outperforms that under the conventional SL framework for both
scenarios: For the static training scenario, the proposed SSL framework shows a
significant improvement in terms of median 2D error, where the median 2D errors
of the proposed SSL framework are closer to the minimum 2D errors of the SL
framework; the results are consistent with those based on the UJIIndoorLoc
database presented in Section~\ref{sec:static-training}. The proposed SSL
framework also significantly reduces IQRs during Periods~0 and 1 compared to
the conventional SL framework, which indicates its robustness for the dynamic
training scenario.

\section{Discussions}
\subsection{Effectiveness Analysis of Mean Teacher}
Wi-Fi RSSI fingerprints exhibit inherent characteristics of high noise, weak
spatial correlation, and spatiotemporal fluctuation. The consistency
regularization strategy of the Mean Teacher enables the model to learn the
robust potential features of RSSI fingerprints through the interaction between
teacher and student models, effectively suppressing noise interference. As can
be seen from the experimental results, the proposed SSL framework can not only
reduce the 3D error but also results in smaller confidence intervals compared to
the conventional SL framework, which demonstrates the effectiveness of the
consistency regularization strategy in enhancing the robustness and
generalization of the localization model, especially in dynamic training
scenarios where the wireless environment is subject to changes.

The AP selection component of the proposed SSL framework helps to align the
feature space of the student and teacher models by selecting APs with more
stable and consistent signal characteristics during each period, which is
crucial for the effectiveness of the consistency regularization. The batch-level
noise injection further enhances the robustness of the model by enabling it to
learn the invariant features that are less sensitive to noise, which is
particularly beneficial in the context of Wi-Fi RSSI data. The interaction
between the teacher and student models allows for a more stable learning
process, as the teacher model provides smoother target labels that can guide the
student model towards better generalization, especially in scenarios with
spatiotemporal variations in RSSI. Moreover, when encountering scenarios where
the entire training dataset is limited, the batch-level noise injection can help
to mitigate overfitting by introducing variability in the training data, which
encourages the model to learn more generalizable features. Overall, these
components work synergistically to improve the performance of the proposed SSL
framework compared to traditional SL approaches and existing DNN models.

\subsection{Research Limitations and Future Directions}
Through the experimental results and their analyses presented in
Section~\ref{sec:exp-results}, the following limitations of the current work
have been identified:
\begin{enumerate}
  \item The proposed AP selection algorithm is mainly designed to meet the feature
        consistency requirement of the SSL framework and does not take into account
        additional factors such as AP load and signal-to-noise ratio (SNR). Also, it
        could be extended to a more adaptive AP selection strategy to systematically
        find an optimal value for the threshold $\tau$ for different datasets and
        scenarios.
  \item The UJIIndoorLoc database lacks real temporal dynamics, resulting in modest
        performance improvement in dynamic training scenarios; the XJTLU dynamic
        database, on the other hand, shows the effectiveness of the proposed SSL
        framework in real dynamic wireless environments, though it is still under
        development and not publicly released yet.
  \item The proposed SSL framework does not currently account for RSSI deviations
        caused by multi-device heterogeneity.
\end{enumerate}

The limitations identified above can be addressed by the following future
research directions:
\begin{enumerate}
  \item Designing a multi-dimensional AP selection algorithm suitable for the SSL
        framework by integrating key features such as AP SNR and load;
  \item generating high-fidelity synthetic unlabeled data based on multi-output
        Gaussian processes to balance the spatiotemporal distribution of the
        dataset~\cite{rela_rnn_02} and constructing more comprehensive dynamic datasets
        covering multiple buildings and device types;
  \item extending the proposed SSL framework to large-scale indoor localization
        scenarios with multi-device heterogeneity.
\end{enumerate}

\section{Conclusions}
\label{sec:con}
In this paper, we have proposed a novel SSL framework for scalable indoor
localization that can exploit unlabeled RSSI fingerprints as well as labeled
ones during both offline and online phases of indoor localization
services. Based on the Mean Teacher model~\cite{ssl:mean_teacher}, the proposed
SSL framework integrates AP selection, pre-training/cloning, and batch-level
noise injection techniques in order to not only expedite SSL training by
mitigating the cold start problem but also enhance robustness and generalization
through consistency regularization.

We conducted extensive experiments to evaluate the localization performance of
reference DNN models under the proposed SSL framework in comparison to that
under the conventional SL framework. The experimental results demonstrate that
the proposed SSL framework consistently improves the indoor localization
performance of the reference models across all the experimental cases, covering
both static and dynamic training scenarios based on the XJTLU dynamic database
as well as the UJIIndoorLoc database. Specifically, for the experiments based on
the UJIIndoorLoc multi-building and multi-floor database, the proposed SSL
framework achieves relative improvements in mean 3D error over the conventional
SL framework up to 7.403\% and 7.748\% for CNNLoc and SIMO-DNN models,
respectively, for the scenario of static training based on a hybrid labeled and
unlabeled database; for the scenario of dynamic training during the online
phase, the proposed SSL framework also provides relative improvements in maximum
3D error of 8.532\% and 12.297\% during Period~1 of the online phase for CNNLoc
and SIMO-DNN, respectively. The results based on the part of the XJTLU dynamic
database covering a single floor, confirm the effectiveness of the proposed SSL
framework with relative improvements in maximum 2D error up to 49.227\% and
39.755\% during Period~0 and Period~1 of the online phase, respectively, under
the dynamic training scenario.

The practical significance of the proposed SSL framework can be highlighted as
follows:
\begin{description}
\item[Reduction in Labeled Data Collection/Maintenance Costs] The proposed SSL
  framework can greatly reduce labeled data collection/maintenance costs while
  maintaining accuracy; the experimental results presented in
  Section~\ref{sec:static-training} suggest that the reduction could be 50\% or
  even more, as SSL with 50\% labeled data outperforms SL with 100\%.
\item[Feasibility of On-Device or Edge Deployment] The computational complexity
  of the proposed SSL framework related with the use of dual models of a teacher
  and a student would not affect target devices because only the teacher model
  needs to be deployed after the training. This results in a smaller parameter
  footprint, which makes its deployment on resource-constrained devices
  feasible. Also, note that, as a training strategy, the proposed framework is
  not limited to models with specific architectures and sizes unlike those based
  on GANs~\cite{rela_SSLComp} and BERT~\cite{rela_BERT}, it can be utilized to
  improve the performance of small models based on simple architectures like the
  simple DNN reference model adopted in Section~\ref{sec:exp-xjtlu-dynamic}.
\item[Support for Continuous Online Adaptation] The proposed SSL framework can
  enhance indoor localization performance of a model over a longer period than
  the conventional SL framework, especially for dynamic wireless environments
  where a localization model can be continuously retrained based on
  newly-submitted unlabeled data from the users of the indoor localization
  system even during the online phase and thereby mitigate performance
  degradation due to environmental changes.
\end{description}

As our proposal is a comprehensive SSL framework integrating the component
techniques of AP selection, pre-training/cloning, and batch-level noise
injection based on the Mean Teacher, each of the component techniques as well
as the underlying SSL method can be extended based on more advanced ones. One
of the most promising among those extensions would be the generation of
synthetic unlabeled data based on more advanced data augmentation techniques
like those based on multi-output Gaussian processes~\cite{rela_rnn_02}, which
could generate labeled data as well for balancing the data distributions (i.e.,
spatial, temporal, or spatiotemporal) in SSL training.




\begin{thebibliography}{10}
\expandafter\ifx\csname url\endcsname\relax
  \def\url#1{\texttt{#1}}\fi
\expandafter\ifx\csname urlprefix\endcsname\relax\def\urlprefix{URL }\fi
\expandafter\ifx\csname href\endcsname\relax
  \def\href#1#2{#2} \def\path#1{#1}\fi

\bibitem{intor_01_1}
A.~Basiri, E.~S. Lohan, T.~Moore, A.~Winstanley, P.~Peltola, C.~Hill,
  P.~Amirian, P.~{Figueiredo e Silva}, Indoor location based services
  challenges, requirements and usability of current solutions, Computer Science
  Review 24 (2017) 1--12.

\bibitem{intor_01_2}
R.~S. Naser, M.~C. Lam, F.~Qamar, B.~B. Zaidan, Smartphone-based indoor
  localization systems: A systematic literature review, Electronics 12~(8)
  (2023).

\bibitem{intro_02}
P.~Bahl, V.~N. Padmanabhan, {RADAR}: An in-building {RF}-based user location
  and tracking system, in: Proc. {INFOCOM}, Vol.~2, Tel Aviv, Israel, 2000, pp.
  775--784.

\bibitem{intro:survey_03}
C.~Basri, A.~El~Khadimi, Survey on indoor localization system and recent
  advances of {WIFI} fingerprinting technique, in: Proc. {ICMCS}, Marrakech,
  Morocco, 2016, pp. 253--259.

\bibitem{intro:survey_04}
C.~Laoudias, A.~Moreira, S.~Kim, S.~Lee, L.~Wirola, C.~Fischione, A survey of
  enabling technologies for network localization, tracking, and navigation,
  {IEEE} Communications Surveys and Tutorials 20~(4) (2018) 3607--3644.

\bibitem{intro:survey_05}
P.~Prasithsangaree, P.~Krishnamurthy, P.~Chrysanthis, On indoor position
  location with wireless {LANs}, in: Proc. {IEEE} {ISPIMRC}, Vol.~2, Lisbon,
  Portugal, 2002, pp. 720--724 vol.2.

\bibitem{knn_01}
J.~Torres-Sospedra, R.~Montoliu, S.~Trilles, \'{O}scar Belmonte, J.~Huerta,
  Comprehensive analysis of distance and similarity measures for {Wi-Fi}
  fingerprinting indoor positioning systems, Expert Systems with Applications
  42~(23) (2015) 9263--9278.

\bibitem{knn_02}
T.~Cover, P.~Hart, Nearest neighbor pattern classification, {IEEE} Transactions
  on Information Theory 13~(1) (1967) 21--27.

\bibitem{Kim:18-1}
K.~S. Kim, S.~Lee, K.~Huang, A scalable deep neural network architecture for
  multi-building and multi-floor indoor localization based on {Wi-Fi}
  fingerprinting, Big Data Analytics 3~(4) (Apr. 2018).

\bibitem{Kim:18-3}
K.~S. Kim, Hybrid building/floor classification and location coordinates
  regression using a single-input and multi-output deep neural network for
  large-scale indoor localization based on {Wi-Fi} fingerprinting, in: Proc.
  {CANDAR}, Hida Takayama, Japan, 2018, pp. 196--201.

\bibitem{cnn_01}
X.~{Song}, X.~{Fan}, X.~{He}, C.~{Xiang}, Q.~{Ye}, X.~{Huang}, G.~{Fang}, L.~L.
  {Chen}, J.~{Qin}, Z.~{Wang}, {CNNLoc}: Deep-learning based indoor
  localization with {WiFi} fingerprinting, in: Proc.
  SmartWorld/{SCALCOM}/{UIC}/{ATC}/{CBDC}om/{IOP}/{SCI}, Leicester, UK, 2019,
  pp. 589--595.

\bibitem{ssl:mean_teacher}
A.~Tarvainen, H.~Valpola, Mean teachers are better role models: Weight-averaged
  consistency targets improve semi-supervised deep learning results (2017).

\bibitem{ssl:overview1}
Y.~Ouali, C.~Hudelot, M.~Tami, An overview of deep semi-supervised learning,
  arXiv preprint arXiv:2006.05278 (2020).

\bibitem{ssl:temporal_ensembling}
S.~Laine, T.~Aila, Temporal ensembling for semi-supervised learning (2016).

\bibitem{rela_GSSL}
L.~Zhang, S.~Valaee, Y.~Xu, L.~Ma, F.~Vedadi, Graph-based semi-supervised
  learning for indoor localization using crowdsourced data, Applied Sciences
  7~(5) (2017).

\bibitem{rela_TSLSSL}
J.~Yoo, Time-series laplacian semi-supervised learning for indoor localization,
  Sensors 19~(18) (2019).

\bibitem{rela_MT_CIR}
P.~Chen, Y.~Liu, W.~Li, J.~Wang, J.~Wang, B.~Yang, G.~Feng, Semi-supervised
  learning-enhanced fingerprint indoor positioning by exploiting an adapted
  mean teacher model, Electronics 13~(2) (2024).

\bibitem{rela_SSLComp}
K.~M. Chen, R.~Y. Chang, {A Comparative Study of Deep-Learning-Based
  Semi-Supervised Device-Free Indoor Localization}, in: Proc.{GLOBECOM},
  Madrid, Spain, 2021, pp. 1--6.

\bibitem{rela_WGAN}
W.~Njima, A.~Bazzi, M.~Chafii, {DNN}-based indoor localization under limited
  dataset using {GAN}s and semi-supervised learning, {IEEE} Access 10 (2022)
  69896--69909.

\bibitem{rela_WePos}
B.~Guo, W.~Zuo, S.~Wang, W.~Lyu, Z.~Hong, Y.~Ding, T.~He, D.~Zhang, {WePos:
  Weak-supervised Indoor Positioning with Unlabeled WiFi for On-demand
  Delivery}, in: Proc. {ACM} {IMWUT}, Association for Computing Machinery, New
  York, NY, USA, 2022, pp. 54:1--54:25.

\bibitem{rela_MTLoc}
J.~Wang, Z.~Zhao, M.~Ou, J.~Cui, B.~Wu, {Automatic Update for Wi-Fi
  Fingerprinting Indoor Localization via Multi-Target Domain Adaptation}, in:
  Proc. {ACM} {IMWUT}, Association for Computing Machinery, New York, NY, USA,
  2023, pp. 78:1--78:27.

\bibitem{rela_BERT}
J.~Devlin, M.~Chang, K.~Lee, K.~Toutanova, {BERT:} pre-training of deep
  bidirectional transformers for language understanding, CoRR abs/1810.04805
  (2018).
\newblock \href {http://arxiv.org/abs/1810.04805} {\path{arXiv:1810.04805}}.

\bibitem{AP_SLC_02}
P.~Roy, M.~Kundu, C.~Chowdhury, Indoor localization using stable set of
  wireless access points subject to varying granularity levels, in: Proc.
  {WiSPNET}, Chennai, India, 2019, pp. 491--496.

\bibitem{AP_SLC_03}
B.~Jia, B.~Huang, H.~Gao, W.~Li, L.~Hao, Selecting critical {WiFi} {AP}s for
  indoor localization based on a theoretical error analysis, {IEEE} Access PP
  (2019) 1--1.

\bibitem{AP_SLC_04}
N.~Saccomanno, A.~Brunello, A.~Montanari, {Let's Forget About Exact Signal
  Strength: Indoor Positioning based on Access Point Ranking and Recurrent
  Neural Networks}, in: Proc. 17th {MobiQuitous}, MobiQuitous '20, Association
  for Computing Machinery, New York, NY, USA, 2021, pp. 215--224.

\bibitem{AP_SLC_05}
M.~Zhou, Y.~Li, M.~J. Tahir, X.~Geng, Y.~Wang, W.~He, Integrated statistical
  test of signal distributions and access point contributions for {Wi-Fi}
  indoor localization, {IEEE} Transactions on Vehicular Technology 70~(5)
  (2021) 5057--5070.

\bibitem{AP_SLC_01}
S.~Li, Z.~Tang, K.~S. Kim, J.~S. Smith, On the use and construction of {Wi-Fi}
  fingerprint databases for large-scale multi-building and multi-floor indoor
  localization: A case study of the {UJIIndoorLoc} database, Sensors 24~(12)
  (2024) 3827.

\bibitem{ColdStart_01}
J.~B. Schafer, D.~Frankowski, J.~Herlocker, S.~Sen, Collaborative Filtering
  Recommender Systems, Springer Berlin Heidelberg, Berlin, Heidelberg, 2007,
  Ch.~9, pp. 291--324.

\bibitem{ColdStart_02}
I.~Turc, M.~Chang, K.~Lee, K.~Toutanova, Well-read students learn better: The
  impact of student initialization on knowledge distillation, {CoRR}
  abs/1908.08962 (2019).
\newblock \href {http://arxiv.org/abs/1908.08962} {\path{arXiv:1908.08962}}.

\bibitem{aug_inject_01}
M.~Eren~Akbiyik, Data augmentation in training {CNNs}: Injecting noise to
  images, arXiv e-prints (2023) arXiv--2307.

\bibitem{aug_inject_02}
M.~Momeny, A.~A. Neshat, M.~A. Hussain, S.~Kia, M.~Marhamati, A.~Jahanbakhshi,
  G.~Hamarneh, Learning-to-augment strategy using noisy and denoised data:
  Improving generalizability of deep {CNN} for the detection of {COVID}-19 in
  {X}-ray images, Computers in Biology and Medicine 136 (2021) 104704.

\bibitem{aug_inject_03}
S.~Yin, C.~Liu, Z.~Zhang, Y.~Lin, D.~Wang, J.~Tejedor, T.~F. Zheng, Y.~Li,
  Noisy training for deep neural networks in speech recognition, {EURASIP}
  Journal on Audio, Speech, and Music Processing 2015 (2015).

\bibitem{misra82:_findin}
J.~Misra, D.~Gries, Finding repeated elements, Science of Computer Programming
  2~(2) (1982) 143--152.

\bibitem{yang25:_unlab_insig_label_boost}
J.~Yang, M.~Chen, Q.~Jia, S.~Liu, Unlabeled insight, labeled boost: Contrastive
  learning and class-adaptive pseudo-labeling for semi-supervised medical image
  classification, Entropy 27~(10) (2025) Art. no. 1015.

\bibitem{arXiv:1909.01804}
Z.~Ke, D.~Wang, Q.~Yan, J.~Ren, R.~W. Lau, Dual student: Breaking the limits of
  the teacher in semi-supervised learning, Arxiv e-prints (2019) 1--11\href
  {http://arxiv.org/abs/1909.01804} {\path{arXiv:1909.01804}}.

\bibitem{ZHANG2025107882}
G.~Zhang, J.~Lu, Y.~Chen, Y.~Deng, B.~Zhao, H.~Chen, L.~Xue, A lightweight
  dual-student mean teacher semi-supervised semantic segmentation method for
  skin lesions, Neural Networks 192 (2025) 107882.

\bibitem{Data:UJI}
J.~Torres-Sospedra, R.~Montoliu, A.~Mart\'inez-Us\'o, J.~P. Avariento, T.~J.
  Arnau, M.~Benedito-Bordonau, J.~Huerta, {UJIIndoorLoc}: A new multi-building
  and multi-floor database for {WLAN} fingerprint-based indoor localization
  problems, in: Proc. {IPIN}, Busan, Korea, 2014, pp. 261--270.

\bibitem{EvAAL}
A.~Moreira, M.~J. Nicolau, F.~Meneses, A.~Costa, {Wi-Fi} fingerprinting in the
  real world -- {RTLS\@UM} at the {EvAAL} competition, in: Proc. {IPIN}, Banff,
  Alberta, Canada, 2015, pp. 1--10.

\bibitem{static_dynamic_training}
Google,
  \href{https://developers.google.com/machine-learning/crash-course/production-ml-systems/static-vs-dynamic-training?hl=zh-cn}{Production
  {ML} systems: Static versus dynamic training} (2024).
\newline\urlprefix\url{https://developers.google.com/machine-learning/crash-course/production-ml-systems/static-vs-dynamic-training?hl=zh-cn}

\bibitem{wknn-02}
R.~Berkvens, M.~Weyn, H.~Peremans, Position error and entropy of probabilistic
  {Wi-Fi} fingerprinting in the {UJIIndoorLoc} dataset, Proc. {IPIN} 2016
  (2016) 1--6.

\bibitem{wknn-KMeans}
S.~Liu, R.~DE~LACERDA, J.~FIORINA, {WKNN} indoor {Wi-Fi} localization method
  using $k$-means clustering based radio mapping, in: Proc. {VTC2021-Spring},
  Helsinki, Finland, 2021, pp. 1--5.

\bibitem{rela_rnn_01}
A.~E. Ahmed~Elesawi, K.~S. Kim, Hierarchical multi-building and multi-floor
  indoor localization based on recurrent neural networks, in: Proc. {CANDARW},
  Matsue, Japan, 2021, pp. 193--196.

\bibitem{rela_rnn_02}
Z.~Tang, S.~Li, K.~S. Kim, J.~S. Smith, Multi-dimensional {Wi-Fi} received
  signal strength indicator data augmentation based on multi-output {Gaussian}
  process for large-scale indoor localization, Sensors 24~(3) (2024).

\bibitem{DumbLoc}
S.~C. Narasimman, A.~Alphones, {DumbLoc: Dumb Indoor Localization Framework
  Using Wi-Fi Fingerprinting}, IEEE Sensors Journal 24~(9) (2024) 14623--14630.

\bibitem{wGan}
W.~Njima, M.~Chafii, R.~M. Shubair, {GAN} based data augmentation for indoor
  localization using labeled and unlabeled data, in: {Proc. {BalkanCom}}, Novi
  Sad, Serbia, 2021, pp. 36--39.

\bibitem{SE-Loc}
Q.~Ye, X.~Fan, H.~Bie, D.~Puthal, T.~Wu, X.~Song, G.~Fang, {SE-Loc:
  Security-Enhanced Indoor Localization With Semi-Supervised Deep Learning},
  IEEE Transactions on Network Science and Engineering 10~(5) (2023)
  2964--2977.
\newblock \href {https://doi.org/10.1109/TNSE.2022.3174674}
  {\path{doi:10.1109/TNSE.2022.3174674}}.

\bibitem{dynamic_static}
Z.~Tang, R.~Gu, S.~Li, K.~S. Kim, J.~S. Smith, Static vs. dynamic databases for
  indoor localization based on {Wi-Fi} fingerprinting: A discussion from a data
  perspective, in: Proc. {ICAIIC}, IEEE, Osaka, Japan, 2024, pp. 760--765.

\end{thebibliography}

\end{document}